%% file: main.tex
\documentclass[10pt,twocolumn,letterpaper]{article}

\usepackage{cvpr}              % To produce the CAMERA-READY version
\input{preamble}
\definecolor{cvprblue}{rgb}{0.21,0.49,0.74}
\usepackage[pagebackref,breaklinks,colorlinks,allcolors=cvprblue]{hyperref}

\usepackage{xcolor}      % 用于定义颜色
\usepackage{tcolorbox}   % 核心包：创建带样式的盒子
\tcbuselibrary{breakable}
\definecolor{noteboxpurple}{RGB}{76, 26, 139}
\definecolor{noteboxbg}{RGB}{240, 235, 250}
\tcbuselibrary{skins}

\newtcolorbox{notebox}[1][]{
    colframe=noteboxpurple,
    colback=noteboxbg,
    coltitle=white,
    colbacktitle=noteboxpurple,
    fonttitle=\small\bfseries,   % 标题字体
    fontupper=\small,            % 正文字体小一号
    boxrule=1.2pt,
    arc=4pt,
    left=10pt, right=10pt, top=8pt, bottom=8pt,
    before title={\vspace{-0.1pt}},
    after title={\vspace{-0.4pt}},
    #1
}

\definecolor{exampleboxborder}{RGB}{224, 182, 87}  % 边框 & 标题栏背景
\definecolor{exampleboxbg}{RGB}{255, 250, 235}     % 内容区背景

\newtcolorbox{examplebox}[1][]{
    breakable,                           % 支持跨页（可选）
    colframe=exampleboxborder,           % 边框颜色
    colback=exampleboxbg,                % 内容区背景
    coltitle=white,                      % 标题文字颜色
    colbacktitle=exampleboxborder,       % 标题栏背景颜色
    fonttitle=\small\bfseries,           % 标题字体小号加粗
    fontupper=\small,                    % 正文字体小号
    boxrule=1pt,                          % 边框粗细
    arc=6pt,                              % 圆角半径（适中）
    left=10pt, right=10pt, top=8pt, bottom=8pt,
    before title={\vspace{-0.1pt}},
    after title={\vspace{-0.4pt}},
    #1                                    % 可选参数，例如 title=...
}

\def\paperID{4482} % *** Enter the Paper ID here
\def\confName{CVPR}
\def\confYear{2026}
\usepackage{multirow}
\def\name{SciEducator}
\def\dataset{SciVBench}

\def\ie{\emph{i.e., }}

\title{\name: Scientific Video Understanding and Educating via Deming-Cycle Multi-Agent System}

\author{ Zhiyu Xu$^1$, 
Weilong Yan$^2$, 
Yufei Shi$^3$, 
Xin Meng$^4$, 
Tao He$^5$, \\
Huiping Zhuang$^6$,
Ming Li$^{7,*}$, 
Hehe Fan$^8$
\\
\\
\ \ \  $^1$Jinan University\quad 
$^2$National University of Singapore\quad 
$^3$Nanyang Technological University\\
$^4$Peking University\quad 
$^5$University of Electronic Science and Technology of China\quad \\
$^6$South China University of Technology
\ \  $^7$Guangming Laboratory\quad 
$^8$Zhejiang University
}

\begin{document}
\maketitle
\input{sec/0_abstract}    
\input{sec/1_intro}

\input{sec/2_formatting}
\input{sec/3_finalcopy}
{
    \small
    \bibliographystyle{ieeenat_fullname}
    \nocite{*}
    \bibliography{main}
}

\input{sec/X_suppl}

\end{document}

%% file: sec/0_abstract.tex
\begin{abstract}
Recent advancements in multimodal large language models (MLLMs) and video agent systems have significantly improved general video understanding. However, when applied to \emph{scientific video understanding and educating}—a domain that demands external professional knowledge integration and rigorous step-wise reasoning—existing approaches often struggle. To bridge this gap, we propose \name{}, the first iterative self-evolving multi-agent system for scientific video comprehension and education. Rooted in the classical Deming Cycle from management science, our design reformulates its \emph{Plan–Do–Study–Act} philosophy into a self-evolving reasoning and feedback mechanism, which facilitates the interpretation of intricate scientific activities in videos. Moreover, \name{} can produce multimodal educational content tailored to specific scientific processes, including textual instructions, visual guides, audio narrations, and interactive references. 
To support evaluation, we construct \dataset{}, a benchmark consisting of 500 expert-verified and literature-grounded science QA pairs across five categories, covering physical, chemical, and everyday phenomena.
Extensive experiments demonstrate that \name{} substantially outperforms leading closed-source MLLMs (e.g., Gemini, GPT-4o) and state-of-the-art video agents on the benchmark, establishing a new paradigm for the community.
\end{abstract}

%% file: sec/1_intro.tex
\section{Introduction}
\label{sec:intro}
%figure1-v3
\begin{figure}[t]
    \centering
    \includegraphics[width=1.0\linewidth]{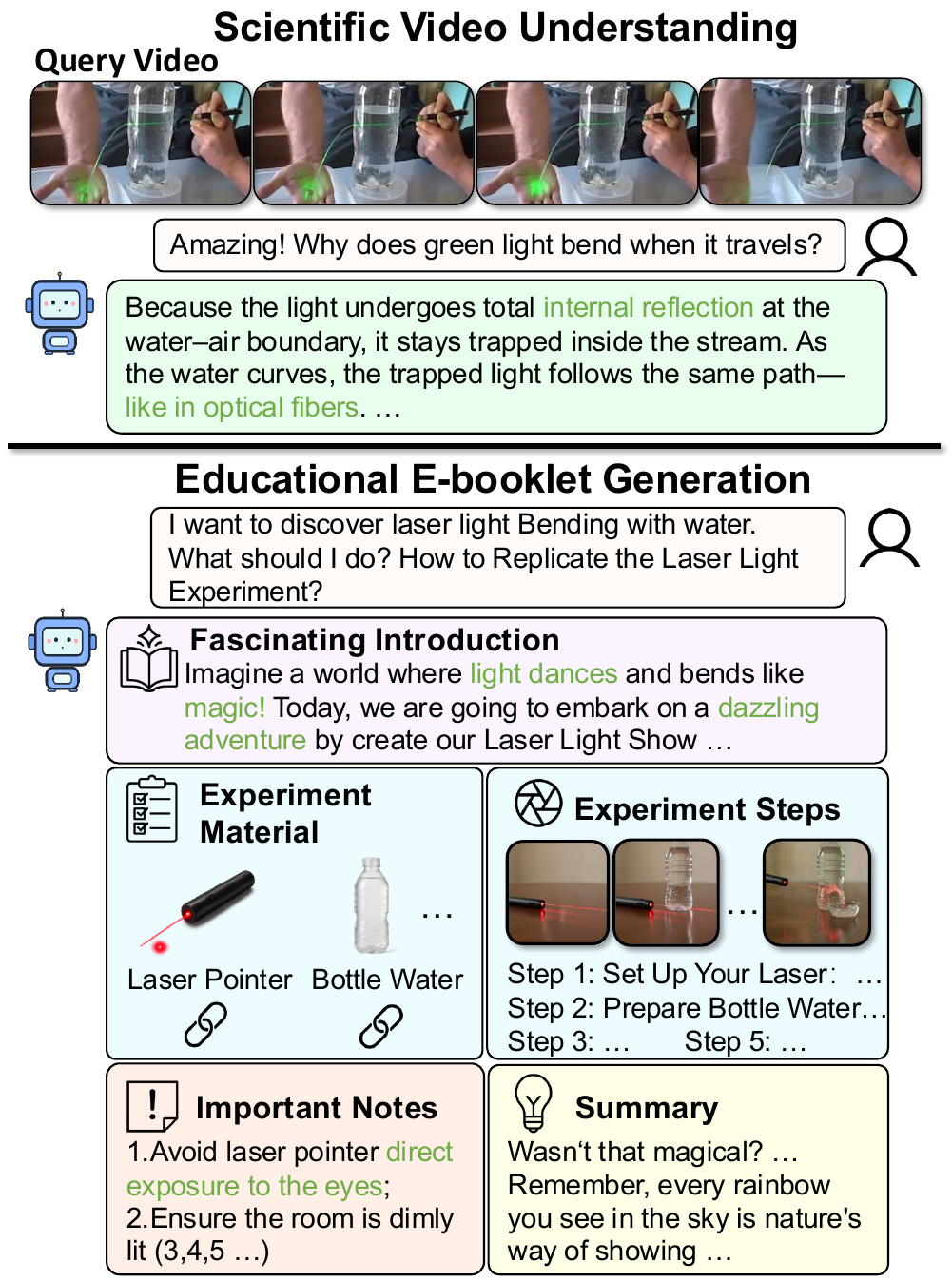}
     \vspace{-0.2cm}
    \caption{SciEducator, conducting video comprehension and delivering science education, can generate multimodal educational e‑booklets that provide comprehensive, detailed, and engaging guidance.}
    \label{fig:Figure 1}
    \vspace{-0.7cm}
\end{figure}

% Video understanding has long been a central research topic in computer vision, with wide-ranging applications across various domains[]. Despite remarkable progress, existing multimodal large language models (MLLMs) and multi-agent systems for video understanding still face substantial challenges. Their comprehension of complex events is constrained by limited internal knowledge, making it difficult to handle queries that require specialized expertise—such as explaining the operational principles of machinery or the causes of specific visual phenomena. Moreover, due to the lack of effective design and guidance, current systems fail to fully exploit the planning, reasoning, and adaptive capabilities of large language models. When confronted with tasks that demand intricate planning or lack straightforward initial solutions, they often struggle to respond effectively. Popular multi-agent video understanding systems can only perform shallow refinements, such as step adjustment or execution retries, without the ability to fundamentally revise their problem-solving strategies[]. Furthermore, most existing architectures lack mechanisms for post-understanding functional development and do not integrate tools to extend their capabilities using multimodal information derived from video comprehension, which severely limits their applicability.

Existing MLLMs for video have achieved significant progress by focusing on enabling large models to understand videos through the integration of visual encoders, large language models, and temporal modeling, thereby allowing them to perceive, reason about, and discuss dynamic visual content~\cite{wang2024internvideo2,xu2024slowfast,cheng2024videollama}. However, they suffer from a key limitation: inadequate ability to leverage external resources and integrate tools~\cite{VideoAgent,videoagent2}, which substantially constrains their functionality and application scenarios.

% Recent advances in agent-based frameworks and tool-augmented reasoning have provided new opportunities to tackle these challenges, particularly for tasks requiring the integration of internal and external knowledge, complex reasoning, and dynamic function expansion. By coordinating multiple LLM-based agents and incorporating diverse tools, modern multi-agent systems can autonomously perform task decomposition, tool invocation, information retrieval, and result synthesis through structured workflows. These systems have achieved impressive performance in complex domains such as software development[] and web search[]. Moreover, by integrating multimodal retrieval and generation modules, they can further enhance video understanding through acquired knowledge, significantly broadening their functional scope and application potential.

Recent agentic systems, capable of integrating and effectively utilizing external tools to ultimately achieve complex and powerful functionalities, thus attracting significant attention~\cite{zhi2025videoagent2,jha2025cross,zhang2025multi}. For instance,~\cite{pang2025paper2poster} proposed a PostAgent that can transform a research paper into a well-structured poster, while~\cite{qian2023communicative} introduced an agentic system integrating multiple agents capable of fully autonomous software development.  These examples demonstrate the substantial application potential of agentic systems. However, they still face several issues at the current stage. Their performance is affected by inherent problems of large language models, such as hallucinations~\cite{ye2023cognitive} and unstable capabilities~\cite{zhao2023survey,laban2025llms,dai2025breach}. Furthermore, for scientific video tasks that require the integration of external professional knowledge and rigorous step-by-step reasoning, these systems often struggle to generate effective and feasible plans initially and still lack the systematic mechanism to self-evolve and self-optimize their workflows based on previous execution results.

To address the limitations of existing agentic systems, we propose \name{}, a novel multi-agent system for scientific video understanding and educating, where external professional knowledge integration, complex task planning, and rigorous reasoning are essential. Rooted in the celebrated \textit{Deming Cycle}~\cite{moen2006evolution,taylor2014systematic} from management science, \name{} introduces an iterative optimization mechanism that enables self-evolving reasoning and continuous enhancement. By repeatedly planning, executing, evaluating, and refining its workflows, the system progressively converges toward high-confidence and accurate interpretations as well as reliable educational content generation of complex scientific activities.

Our \name{} operates in two stages: \textit{understanding} and \textit{educating}. In the \textit{understanding} stage, an LLM-based planner acts as the central controller. It retrieves domain knowledge from both internal and external sources, formulates multiple candidate workflows involving various agents (e.g., Video Content Acquisition Agent, Web or Academic Paper Search  Agent), and stores them in a solution pool. An LLM-based evaluator then integrates diverse metrics to assess each candidate and select the optimal workflow for execution. The execution results undergo a confidence evaluation: if the planner deems the current results sufficiently convincing, it synthesizes the available information to produce the final output; otherwise, \name{} proceeds with further processing. The execution outcomes and feedback, especially summarized failure reasons and newly acquired knowledge, are subsequently used to refine and update the solution pool, thereby embodying the Deming Cycle’s iterative improvement loop. In the \textit{educating} stage, \name{} generates multimodal e-learning materials tailored to the scientific content interpreted in the \textit{understanding} stage. It first retrieves relevant experimental procedures, safety precautions,  required equipment images, and shopping links, and then produces textual instructions, visual guides, audio narrations, and interactive e-booklets that present scientific knowledge in an engaging and accessible manner.

To evaluate the superiority of our system, we introduce \dataset, a new benchmark containing 500 expert-validated and literature-grounded question–answer (QA) pairs across three scientific domains: physics experiments, chemistry experiments, and daily life phenomena. The QA pairs are further categorized into five types, \ie terminology, principle, prediction, reading, and design. We also propose a comprehensive set of metrics for comparative evaluation of both understanding and educating capabilities. Experimental results show that \name{} consistently outperforms leading commercial MLLMs and state-of-the-art multi-agent systems in both aspects.

Our main contributions are summarized as follows:
\begin{itemize}
\item We propose \textbf{\name}, the first multi-agent system for scientific video understanding and educating, which not only comprehends complex and fine-grained scientific videos but also generates multimodal educational e-booklets that integrate diverse information to deliver engaging guidance and stimulate scientific curiosity.
\item Inspired by the \textbf{Deming Cycle}, we equip \name{} with an iterative workflow optimization mechanism that progressively refines solutions through failure analysis and newly acquired knowledge, substantially enhancing its capability to handle meticulous scientific activities and offering a novel paradigm for agentic system design.
\item We construct \textbf{\dataset}, the first benchmark for scientific-phenomenon video analysis, featuring diverse question–answer pairs across physics, chemistry, and daily life phenomena, along with a comprehensive set of evaluation metrics.
\end{itemize}
%-------------------------------------------------------------------------

%% file: sec/2_formatting.tex
\section{Related Work}
\label{sec:body}

\subsection{Multimodal Large Language Models}
Recent years have witnessed substantial progress in vision-language models applied to video understanding. VTimeLLM introduces a boundary-aware three-stage training strategy~\cite{huang2024vtimellm}, TimeSuite employs temporal adaptive position encoding~\cite{zeng2024timesuite}, and Slowfast-LLava designs a low-speed feature sampler to enhance temporal awareness in visual representations~\cite{xu2024slowfast}. Closed-source models such as GPT-4o~\cite{openai_gpt4o_system_card_2024}, Gemini~\cite{google_gemini20_flash_2024}, and Claude~\cite{anthropic_claude37_sonnet_2025} have also been widely adopted as foundational MLLMs in downstream video tasks. However, these systems are critically limited by their inherent inability to effectively leverage external resources and integrate specialized tools. This fundamental shortfall severely curtails their functional capacity and practical applicability in various scenarios.

\subsection{Multi-Agent Systems}

With the widespread application of multi-agent systems(MASs) across numerous domains, computer vision researchers have also begun exploring MASs for video understanding~\cite{yang2024doraemongpt,gao2023assistgpt}. VideoAgent~\cite{VideoAgent} proposed a framework capable of interactive reasoning and planning through confidence evaluation, focusing particularly on long-form video comprehension. In ~\cite{videoagent2}, a memory-enhanced MAS for video understanding was introduced, emphasizing structured spatiotemporal information. VideoAgent2~\cite{zhi2025videoagent2} simulated the cognitive process of human video understanding by designing an uncertain chain-of-thought reasoning approach. However, they still face several challenges. Their performance is limited by the inherent weaknesses of LLM, like hallucinations and instability. For scientific video tasks requiring external expertise and rigorous step-by-step reasoning, they often fail to produce viable initial plans and lack systematic mechanisms that enable them to self-evolve and optimize their workflows based on past results.

\begin{figure*}[ht]
    \centering
    \includegraphics[width=1.0\linewidth,height=9.2cm]{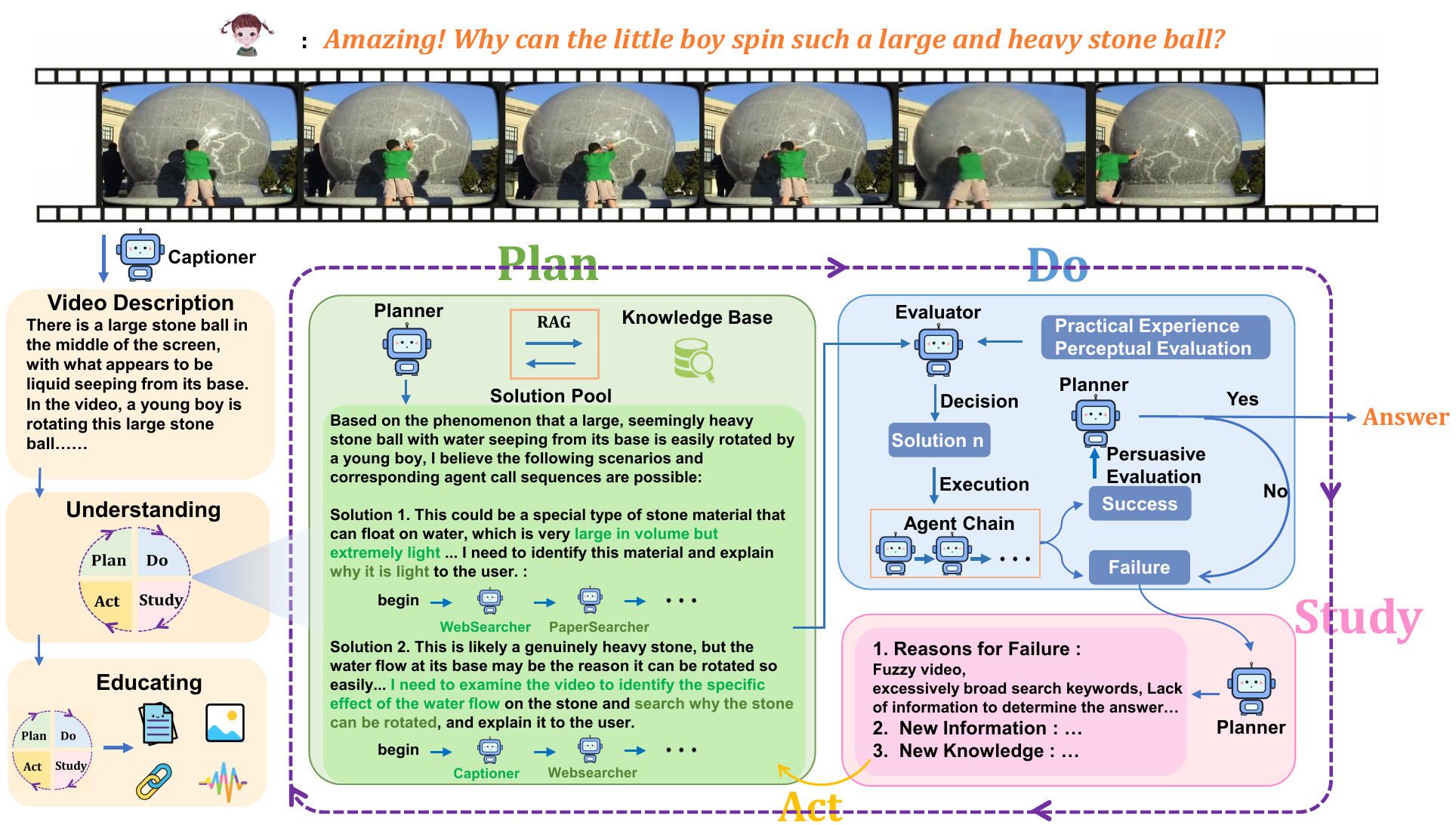}
    \caption{SciEducator architecture overview. We design a multi-agent system capable of implementing the PDSA cycle, which iteratively optimizes output responses through cyclic iterations. }
    \label{fig:pipeline}
    \vspace{-0.5cm} 
\end{figure*}

\section{\name}
 Our pipeline is illustrated in \cref{fig:pipeline}, where we introduce SciEducator, a multi-agent system for scientific video comprehension and education. We begin by giving the problem definition in \cref{sec:3.1Problem Formulation}, and introduce the specialized tools \& agents integrated into the system in \cref{sec:3.2tools}, followed by a description of SciEducator's video understanding workflow in \cref{sec:3.3Deming Cycle}, and the educational booklet generation process in \cref{sec:3.4booklet_generation}. 
 % Finally, we elaborate on the design philosophy, architecture, and operational logic of the PDSA cycle in SciEducator in Section 3.5. 
 These designs enable the system to effectively analyze problems, provide feasible solutions, and ultimately produce accurate and robust responses, with an optional education e-booklet generation.

\subsection{ Problem Definition and  Overview}

\label{sec:3.1Problem Formulation}
% As discussed, we drew inspiration from the Deming Cycle (PDSA) and endowed the system with the ability to iteratively optimize solutions based on prior execution results and newly acquired knowledge. This dynamic process enables the system to progressively converge toward the correct answer. The core workflow of our system can be formally expressed as:

% \begin{equation}
% a = S^n(q;\mathcal{T})
% \end{equation}

% Here, $S$ denotes our multi-agent system, and the notation $S^n$ indicates that the system can iteratively execute this process multiple times depending on specific circumstances. The input question $q$ represents a concrete query related to a scientific video, while $\mathcal{T}$ refers to a dynamically configurable set of tools or agents, expressed as $\mathcal{T} = {t_1, t_2, \dots, t_n}$, where $t_i$ denotes an individual tool or agent capable of performing tasks based on input parameters. Each $t_i \in \mathcal{T}$ executes specific tasks according to its inputs and produces corresponding outputs. The system aggregates outputs from various tools, integrates and reasons over the information, and evaluates whether the currently acquired information is sufficient to answer the user's query and corresponds to the video content—that is, it assesses the confidence level. Based on this confidence level, the system ultimately decides whether to terminate the process and output answer $a$, or to proceed with reprocessing.

The goal of our multi-agent system $\mathcal{S}$ is to return an accurate, self-consistent answer $A$ given a user query $Q$ and a scientific video $V$; what's more, it is optional to generate an educational booklet after video understanding. The core workflow of our approach can be formally represented as follows:
% \begin{align}
% A &= \mathcal{S}\!\big(Q,\, V,\, P,\, E,\, \mathcal{T}\big), \label{eq:system}
% \end{align}
\begin{align}
A = \mathcal{S}(Q, V; P, E, \mathcal{T}),
\label{eq:system}
\end{align}
where $P$ and $E$ represent the planner and evaluator, which act as the leading roles inside the Deming (PDSA) cycle. $\mathcal{T}$ denotes the dynamically configured set of tools and agents aiming at different stages inside our system. For example, in the Deming cycle, $\mathcal{T}$ includes $\mathcal{T}_{\text{Plan}}, \mathcal{T}_{\text{Do}}, \mathcal{T}_{\text{Study}}$; and in the educational booklet generation stage, $\mathcal{T}$ refers to $\mathcal{T}_{\text{Booklet}}$.

Speaking of the PDSA iteration, which is the core of our pipeline, it conducts the plan-do-study-act process iteratively to obtain the most refined and reasonable answer, as shown in \cref{fig:pipeline}. Specifically, "plan" refers to the process of retrieving the corresponding knowledge and building up a solution pool from our system's internal knowledge base. The "do" stage targets at assessing to find the most appropriate solution from the pool, and provides confidence to decide either to stop cycles and give a well-structured response to users, or step into the "study" stage. In "study", our system aims to find the reasons for being unable to give answers, and seeks useful new information. After updating the solution pool with the "act" stage, it will have better confidence to provide a good reply to the query. We will illustrate more details in \cref{sec:3.3Deming Cycle}.

Another key aspect of our system is the extension of scientific video understanding toward educational content generation, enabling the automatic creation of structured e-booklets that illustrate and guide the replication of scientific phenomena with high relevance, clarity, and appeal to young learners. Further details are provided in \cref{sec:3.4booklet_generation}.

% \todo{will move the following to 3.3.} we formally formulate it with each step in the $i^{th}$ cycle below:  \todo{how to organize these formulations?}
% \begin{align}
% M_i &= P\!\big(Q,\, V,\, \mathcal{T}_{\text{Plan}}\big), \label{eq:plan_stage}
% \end{align}
% \begin{align}
% R_i &= E\!\big(M_i,\, V,\, \mathcal{T}_{\text{Do}}\big), \label{eq:do_stage}
% \end{align}
% \begin{align}
% C_i &= P\!\big(R_i,\, Q,\, V\big), \label{eq:confidence_stage}
% \end{align}
% \begin{align}
% F_i, \, K_i &= P\!\big(R_i, \, Q,\, V,\, \mathcal{T}_{\text{Study}}\big), \label{eq:study_stage}
% \end{align}
% \begin{align}
% M_{i+1}&= \Gamma_{\text{Act}}\!\big(F_i,\, K_i,\, M_i\big), \label{eq:act_stage}
% \end{align}
% where $M, R$ refer to the solution pool from the planning stage and the chosen result in the pool at the do stage. $C$ represents the evaluated confidence, which guides the following study process to obtain failure analysis $F$ and knowledge $K$. 
% $\mathcal{T}_{\text{Plan}}, \mathcal{T}_{\text{Do}}, \mathcal{T}_{\text{Study}}$ are the sets of tools and agents corresponding to different stages.
% $\Gamma_{\text{Act}}$ is the act function that provides the solution pool for the next iteration. 
% }

\subsection{Tools \& Agents Configuration}
\label{sec:3.2tools}

Our system integrates a total of sixteen specialized components, including ten agents and six tools, each tailored for specific subtasks with well-defined input–output interfaces. These components are organized into two categories: (i) \textbf{dynamically invocable tools/agents}, which are adaptively called based on the system’s reasoning context; and (ii) \textbf{fixed-execution tools/agents}, which operate automatically at predetermined stages of the workflow.

The dynamically invocable components primarily support functions such as task planning, content acquisition, evaluation, web or literature retrieval, and safety prompting. The fixed-execution components focus on knowledge-base maintenance, multimodal synthesis, and e-booklet generation. Detailed descriptions of all tools and agents, including their configurations and interdependencies, are provided in the Supplementary Material.

\subsection{Deming Cycle within SciEducator}
\label{sec:3.3Deming Cycle}

We instantiate SciEducator as a closed-loop Plan–Do–Study–Act (PDSA) controller tailored to scientific video understanding and its downstream educational content creation. The same loop is later reused for e-booklet generation (\cref{sec:3.4booklet_generation}) with a slightly different toolset and, when applicable, without video input. Each of the four stages in the $i$-th cycle is described as follows.

\paragraph{Plan stage.}
Given a user query $Q$ and video $V$ sampled at one fps, we prompt a captioner to obtain a temporally grounded description $V_{\text{content}}$. A retrieval-augmented agent then extracts salient entities and keywords and retrieves domain knowledge $K$ from an internal corpus. Noted that, this $K$ is only retrieved in the first cycle for once, and will be updated just by the study stage.  The planner composes a diverse pool of candidate plans by combining $V_{\text{content}}$ and $K$, using few-shot exemplars to stabilize call signatures and parameter settings. This stage can be summarized as:
\begin{align}
M_i, K &= P\big(Q,\, V,\, \mathcal{T}_{\text{Plan}}\big),
\label{eq:plan_stage}
\end{align}
where $M_i$ is the candidate plan pool in the $i$-th cycle and $\mathcal{T}_{\text{Plan}}$ denotes the planning toolset.

\paragraph{Do stage.}
Given the solution pool $M_i$, the evaluator scores each plan with a composite of objective $A_{\text{obj}}$ and LLM-based perceptual criteria $A_{\text{percep}}$  and selects the best trade-off in time and token efficiency, success likelihood, feasibility, and overall performance. To estimate time and token costs, we build an empirical prior $\mathcal{E}$ by issuing 20 randomized probe calls per tool/agent, from which we obtain average latency, average token usage, and success probability. For relevance and feasibility, we weight a plan’s keywords using inverse document frequency (IDF) ~\cite{ramos2003using}, reflecting their ability to retrieve discriminative knowledge for the current video:
\begin{align}
\mathrm{IDF}(k)=\log\!\left(\frac{N}{\mathrm{f}(k)+1}\right),
\end{align}
where $N$ is the corpus size and $\mathrm{f}(k)$ is the number of documents containing $k$; we use an internal corpus of 84 physics and chemistry documents. In parallel, a subjective judge compares plans on coverage, logical coherence, scientific soundness, and clarity. The selection of the best solution $s^\star$ is defined as:
\begin{align}
s^\star=\arg\max_{s\in M_i}\big[\,A_{\text{obj}}(s;\mathcal{E},\mathrm{IDF})+\lambda\,A_{\text{percep}}(s)\,\big],
\end{align}
which is then executed step-by-step to produce the stage output $R_i$. We summarize the do stage as
\begin{align}
R_i &= E\big(M_i,\, V,\, \mathcal{T}_{\text{Do}}\big),
\label{eq:do_stage}
\end{align}
where $E(\cdot)$ is the evaluator and  $\mathcal{T}_{\text{Do}}$ refers to the do-stage toolset.
After obtaining $R_i$, the planner estimates a confidence score using the query $Q$, the video context $V$, and the executed plan:
\begin{align}
C_i &= P(R_i,\, Q,\, V).
\label{eq:confidence_stage}
\end{align}
This score indicates whether the available evidence suffices to address the query and yield a convincing answer; if high, SciEducator synthesizes the information and produces a well-structured response.

\paragraph{Study and Act Stage.}
If confidence is low, the system enters the study phase: the planner diagnoses why the answer is insufficient (e.g., tool failures, overly broad and irrelevant retrieval, or insufficient detail in video captions) and aggregates any useful evidence discovered this round into the knowledge base. Formally,
\begin{align}
F_i,\, K_{i+1} &= P\big(R_i,\, K_i,\, Q,\, V,\, \mathcal{T}_{\text{Study}}\big),
\label{eq:study_stage}
\end{align}
where $F_i$ is the failure analysis and $K_{i+1}=K_i\cup K_{\text{new}}$ is the updated knowledge. Guided by $F_i, K_i$, the act function $\Gamma_{\text{Act}}$ replans the next pool—adjusting structure, granularity, or query specificity (e.g., apply video super-resolution for blurry frames, increase captioning fps for missed actions, and refine queries with more specific entities):
\begin{align}
M_{i+1} &= \Gamma_{\text{Act}}\big(F_i,\, K_{i+1},\, M_i\big).
\label{eq:act_stage}
\end{align}
The system executes the next iteration with the updated solution pool $M_{i+1}$ and repeats the confidence check, iterating until $C_i\!\ge\!\tau$ or a maximum number of cycles we choose. 

% Empirically, expected answer quality improves across iterations until saturation:
% \[
% \mathbb{E}_{i+1,Q,V}[a] \;>\; \mathbb{E}_{i,Q,V}[a],
% \]
% where $\mathbb{E}(a)$ denotes the expected answer quality (average relevance and accuracy).

\subsection{Educational E-booklet Generation}
\label{sec:3.4booklet_generation}

After video understanding, SciEducator identifies the specific scientific phenomenon in the video and its underlying principles. When required, it triggers a multimodal retrieval and generation pipeline to produce a child-friendly e-booklet that includes: \textbf{(a)} experimental guidance; \textbf{(b)} required equipment with purchase links; \textbf{(c)} step-by-step procedures with instructional diagrams; \textbf{(d)} safety precautions for both equipment and operations with accompanying audio prompts; and \textbf{(e)} a concise summary of the experiment and its principles. The booklet adopts a progressive, logically organized layout to support rapid comprehension and faithful replication by young readers.

To assemble content, SciEducator uses an Entity Recognition agent to extract key entities, then the Procedure Search and Safety Alert agents to retrieve replicable steps and precautions. We target four properties: \textbf{(i)} relevance (avoiding content unrelated to the experiment), \textbf{(ii)} instructional quality (complete, detailed, and safety-aware), \textbf{(iii)} attractiveness (engaging, visually impressive replication), and \textbf{(iv)} educational value (steps that expose the governing principles). The process follows the same PDSA loop as \cref{sec:3.3Deming Cycle}, but here the input is only the query, confidence is assessed w.r.t.\ these four properties, while the available tools are different.

After obtaining the procedures, an Equipment Search tool returns images and purchase links for the relevant equipment; an Illustration Generation tool produces step-wise guidance images; and a Speech Generation tool synthesizes audio instructions. Finally, an E-booklet Generation agent integrates text, images, audio, hyperlinks, and layout into a well-structured booklet in the style of children’s science literature. Compared with a single-LLM baseline, SciEducator provides richer modalities with higher relevance and instructional quality, while improving engagement through integrated text–image–audio design.

\newcommand{\vsepLR}{\hspace{4pt}}
\makeatletter
% 1) 上下都留白
\newcommand{\VSEPboth}{%
  \vsepLR\vrule width 0.3pt
    height \dimexpr\ht\@arstrutbox-0.35ex\relax
    depth  \dimexpr\dp\@arstrutbox-0.35ex\relax
  \vsepLR
}
% 2) 只上留白（下不留）
\newcommand{\VSEPtop}{%
  \vsepLR\vrule width 0.3pt
    height \dimexpr\ht\@arstrutbox-0.35ex\relax
    depth  \dp\@arstrutbox
  \vsepLR
}
% 3) 只下留白（上不留）
\newcommand{\VSEPbot}{%
  \vsepLR\vrule width 0.3pt
    height \ht\@arstrutbox
    depth  \dimexpr\dp\@arstrutbox-0.35ex\relax
  \vsepLR
}
% 4) 上下都不留白
\newcommand{\VSEPnone}{%
  \vsepLR\vrule width 0.3pt height \ht\@arstrutbox depth \dp\@arstrutbox\vsepLR
}
\makeatother

\section{Experiments}

\subsection{SciVBench Dataset}
\begin{figure}[t]
    \centering
    \includegraphics[width=0.9\linewidth]{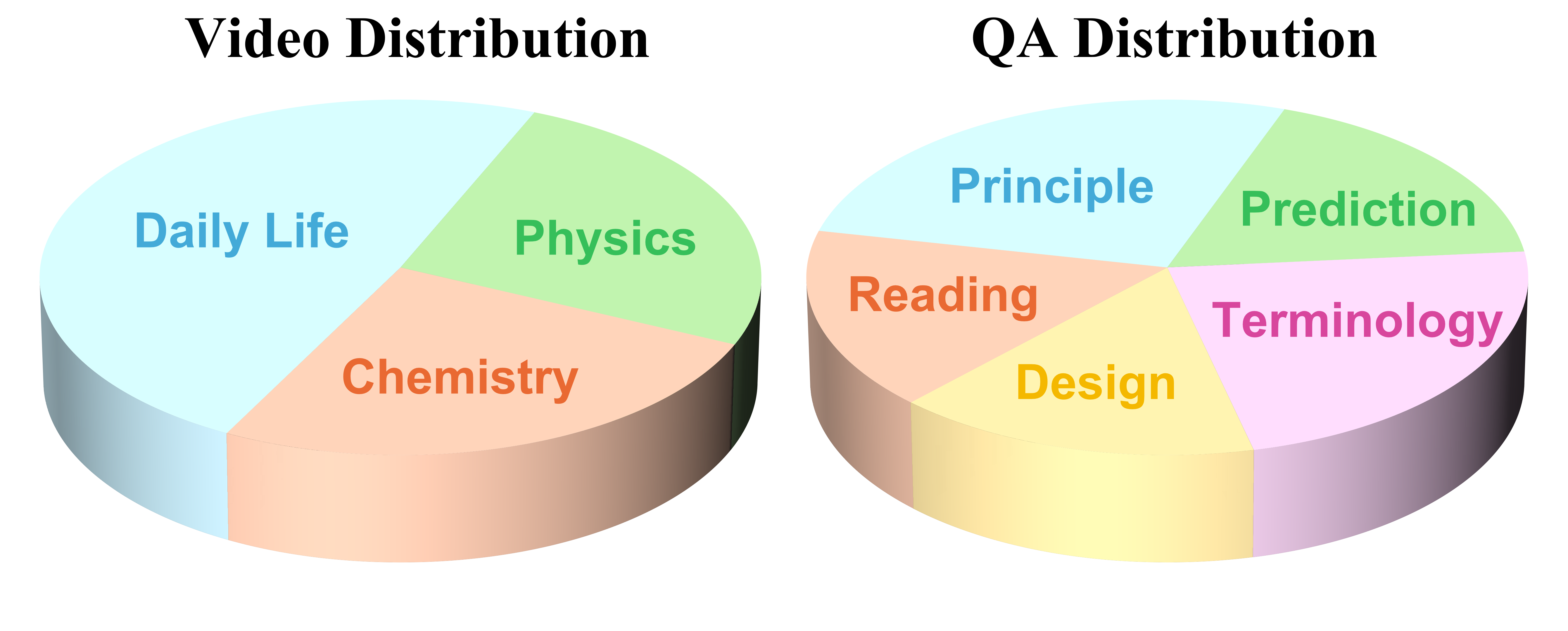}
    \vspace{-0.2cm} 
    \caption{Proportional distribution of video and QA categories. SciVBench comprises three types of scientific videos and five categories of questions, enabling a comprehensive evaluation of a model’s ability to acquire diverse domain knowledge and tackle various complex scientific problems. }
    \label{fig:Figure 2}
    \vspace{-0.5cm} 
\end{figure}

To construct a specialized, accurate, and reasoning-intensive collection of scientific videos with corresponding QA pairs, we systematically gathered rich scientific video resources from major video platforms and science education websites, including $54$ physics experiment videos, $54$ chemistry experiment videos, and $103$ daily life phenomenon videos as shown in \cref{fig:Figure 2}. Domain experts in physics and chemistry systematically reviewed the subtitles, explanatory narrations, knowledge manuals or guides, as well as relevant reference materials and academic papers for each video. Two domain experts independently authored and validated each QA pair, and disagreements were adjudicated by a third expert. Through rigorous cross-validation with additional sources, we developed $500$ scientifically-grounded question-answer pairs based on the video content: $160$, $148$, and $192$ pairs for physics, chemistry, and daily life phenomenon videos. 

Only the visual content of the videos was used as input, with all subtitles and audio narrations removed. Each QA pair was designed such that correct answers could only be derived by comprehending the video content, not from the question text alone. All answers maintain logical integrity, forming coherent reasoning chains from fundamental knowledge to final conclusions, while avoiding any semantic deviation from reference materials and papers. These highly specialized, accurate, and reasoning-intensive QA pairs establish a solid foundation for evaluating the performance of existing MLLMs and multi-agent systems (MASs) in scientific video comprehension and validating the effectiveness of our SciEducator.

% \begin{figure}[t]
%     \centering
%     \includegraphics[width=0.8\linewidth,height=2.8cm]{sec/portionv6.png}
%     \caption{Proportional distribution of video and QA categories. SciVBench comprises three types of scientific videos and five categories of questions, enabling a comprehensive evaluation of a model’s ability to acquire diverse domain knowledge and tackle various complex scientific problems. }
%     \label{fig:Figure 2}
%     \vspace{-0.5cm} 
% \end{figure}

\begin{figure*}[t]
    \centering
    \includegraphics[width=1.0\linewidth]{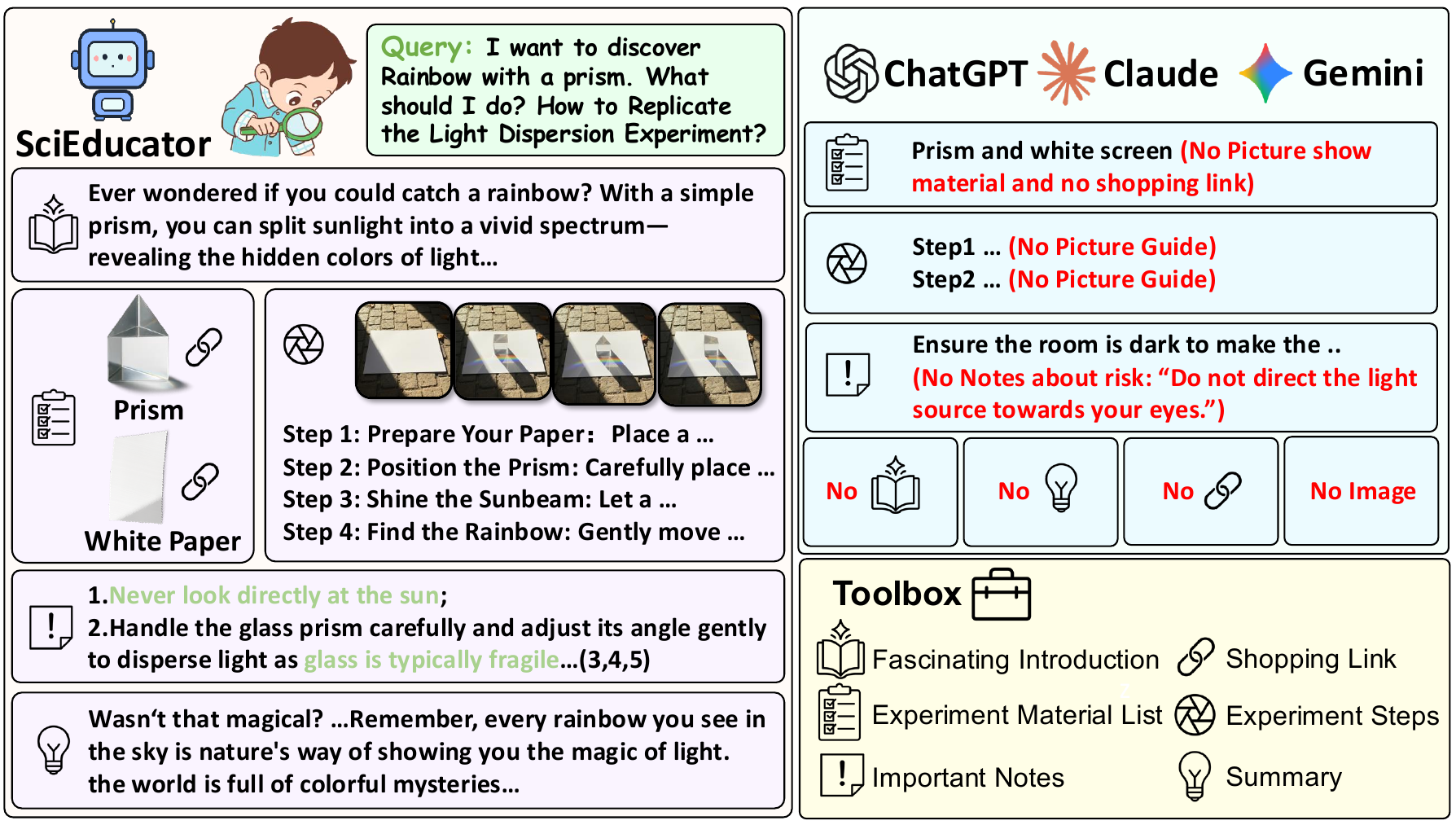}
    \caption{Qualitative comparison between SciEducator and MLLMs in Education E-booklet Generation. The left is our generated e-booklet with comprehensive contents and a well-organized structure, while other popular MLLMs (see the right) all fail in such generation. These examples demonstrate SciEducator's ability to generate more comprehensive, better-structured, and more attractive Education E-booklet.}
    \label{Qualitative with sota education}
    \vspace{-0.1cm}
\end{figure*}

\subsection{Evaluation Metrics}
\paragraph{Evaluation Metrics for Understanding.}
We employ two metrics to evaluate the performance in scientific video understanding, using qwen3-max as the unified evaluator for all model-generated answers. Specifically, these metrics are:
\begin{itemize}
    \item \textbf{Relevance (Rel)}: We first provide the reference answer and extensive scientific background for the current question to Qwen3-Max \cite{bai2023qwen}, instructing it to assess how relevant the model-generated answer is to the query—focusing on whether the response aligns with the scientific subdomain involved, regardless of correctness. The goal is to evaluate whether the answer provides misleading or irrelevant information. The scoring uses three discrete values: $0$ for irrelevant responses, $0.5$ for partially relevant, and $1$ for fully relevant.
    \item \textbf{Accuracy (Acc)}: We then use Qwen3-Max \cite{bai2023qwen} to analyze the semantic similarity between the generated answer and the reference answer, scoring the correctness of the model's response. Similarly, this metric employs three discrete scores: $0$ for completely incorrect answers, $0.5$ for partially correct, and $1$ for fully correct.
\end{itemize}

Detailed scoring strategies and prompts can be found in the supplementary materials. The relevance and accuracy metrics for each model are calculated as the average scores across all questions and are presented in the results table as percentages, with the percentage symbols omitted.

\paragraph{Evaluation Metrics for Education.}
To assess the educational performance of SciEducator and various MLLMs, we decoupled this part from the understanding benchmark. All models were provided with the scientific terminology of the current experiment in a video and asked to generate procedures and precautions for replicating it. Although SciEducator can generate more modalities than MLLMs, we focused specifically on comparing their shared textual modality for quantitative comparison.

\makeatletter
% 仅用于 Max Round 与 Physics 之间的竖线：左右各 1pt；上下各留 0.1ex（更贴近横线）
\newcommand{\ColSep}{%
  \hspace{1pt}\vrule width 0.25pt
    height \dimexpr\ht\@arstrutbox-0.10ex\relax
    depth  \dimexpr\dp\@arstrutbox-0.10ex\relax
  \hspace{1pt}%
}
\makeatother

% \begin{table}[t]
% \centering
% \caption{Quantitative Comparison of SciEducator with different maximum iteration rounds (all rows are SciEducator). Performance rises with larger Max Rounds, visualizing the benefit of iterative PDSA cycles.}
% \scriptsize
% \setlength{\tabcolsep}{6pt}        % 列间距更小
% \renewcommand{\arraystretch}{0.99}   % 行距更紧
% \setlength{\aboverulesep}{0.1ex}
% \setlength{\belowrulesep}{0.1ex}

% \begin{tabular}{@{}c@{\ColSep}*{6}{c}@{}}
% \toprule
% \multirow{2}{*}{Max Round(s)} &
% \multicolumn{2}{c}{Physics} &
% \multicolumn{2}{c}{Chemistry} &
% \multicolumn{2}{c}{Daily Life} \\
% \cmidrule(l{1pt}r{0pt}){2-7} % 连续横线；左端与竖线仅留 1pt
%  & Rel & Acc & Rel & Acc & Rel & Acc \\
% \midrule
% 1 & 56.56 & 42.50 & 49.32 & 40.88 & 36.20 & 33.33 \\
% 2 & 65.63 & 49.38 & 53.04 & 44.26 & 46.88 & 44.27 \\
% 3 & 77.19 & 58.75 & 62.16 & 52.70 & 52.86 & 51.04 \\
% 4 & 80.31 & 63.75 & 73.31 & 63.18 & 62.24 & 59.38 \\
% 5 & \textbf{81.88} & \textbf{65.31} & \textbf{73.97} & \textbf{64.86} & \textbf{64.58} & \textbf{62.24} \\
% \bottomrule
% \end{tabular}
% \label{Quantitative with different maximum iteration rounds}
% \end{table}

We employed four metrics and uniformly used Qwen-VL-Plus \cite{bai2023qwen} to evaluate all model responses in a comparative setting. Qwen-VL-Plus \cite{bai2023qwen} was supplied with substantial background information about each experiment as a reference. Specifically, the metrics are:
\begin{itemize}
    \item \textbf{Relevance}: How well the generated experimental procedures and precautions align with the current experiment and its underlying principles.
    \item \textbf{Instructional Quality (IQ)}: How effectively the generated procedures and precautions guide children in conducting the experiment, with emphasis on detail orientation, completeness, clarity, and safety warnings.
    \item \textbf{Attractiveness}: A comprehensive assessment of how engaging the textual instructions are. For SciEducator, the aesthetic quality of its supporting illustrations is also incorporated into this evaluation to identify the most captivating response.
    \item \textbf{Educational Value (EV)}: How well each model’s response stimulates children’s scientific interest and guides them to understand the principles through the experiment.
\end{itemize}

Detailed scoring strategies and prompts can be found in the supplementary materials. We evaluate $40$ videos (Education Subset). For each video, all model responses were collectively anonymized and input into the VLM for evaluation. The best response for each metric was selected, and final win rates were calculated and presented in tables.

\begin{figure*}[t]
    \centering
    \includegraphics[width=1.0\linewidth]{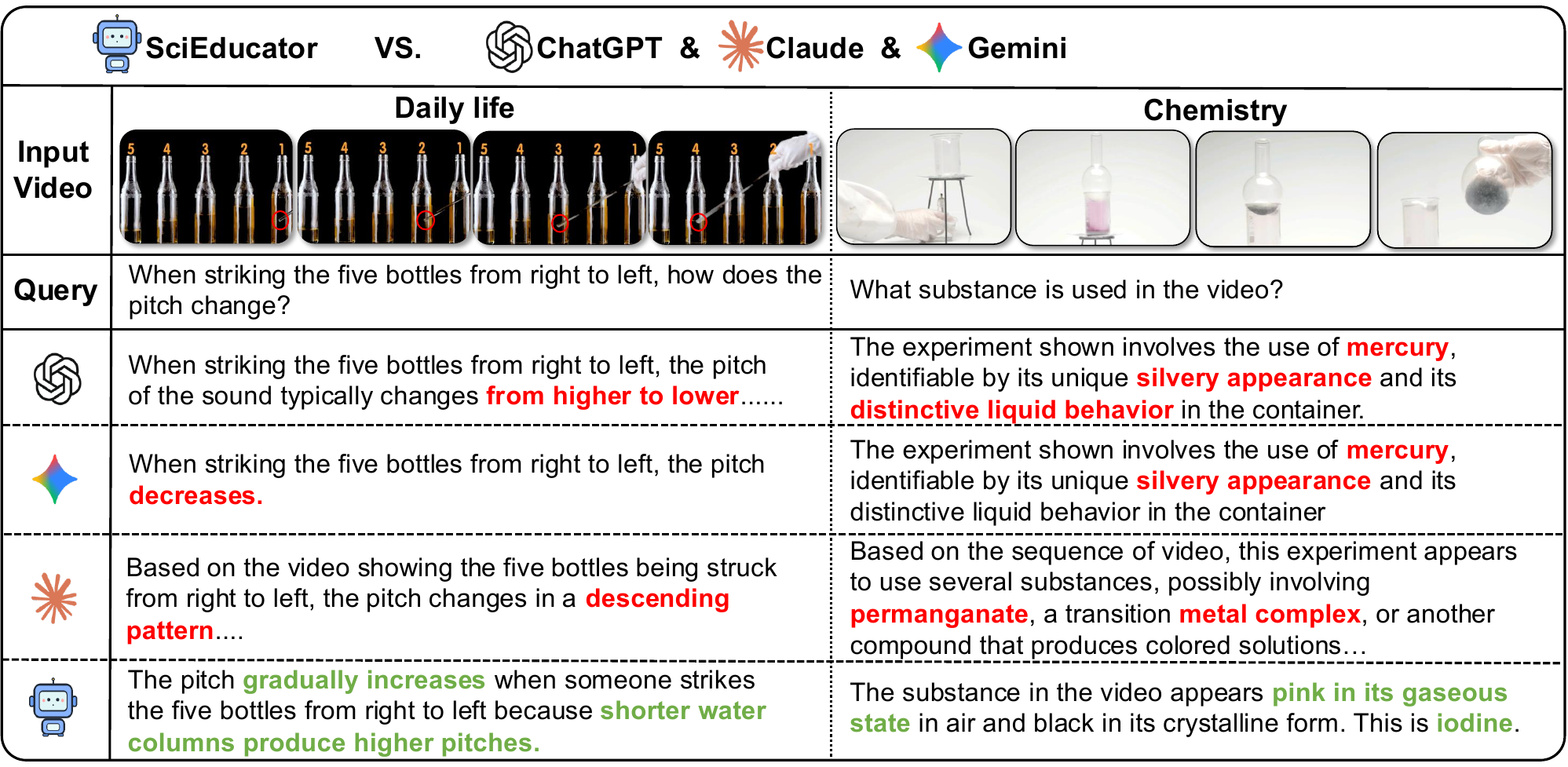}
    \caption{Qualitative comparison between SciEducator and MLLMs. These examples demonstrate SciEducator's ability to generate more comprehensive, better-structured, and more logically coherent answers than the other MLLMs. }
    \label{Qualitative with sota video understanding}
\end{figure*}

\subsection{Experimental Results}
\paragraph{Quantitative Analysis.}

To verify SciEducator's effectiveness, we evaluate SciEducator on SciVBench against three closed‑source MLLMs (Claude 3.7 Sonnet \cite{anthropic_claude37_sonnet_2025}, GPT‑4o \cite{openai_gpt4o_system_card_2024}, Gemini 2.0 Flash \cite{google_gemini20_flash_2024}) and two state-of-the-art multi‑agent systems (MASs) (VideoAgent \cite{VideoAgent} and videoagent \cite{videoagent2}), and report results in \cref{quantitative comparison with sota} for scientific video understanding and \cref{SciEducator with sota} for educational content generation.

% To verify SciEducator's effectiveness, we present the quantitative results about the scientific video understanding ability in \cref{quantitative comparison with sota}, which is compared with two state-of-the-art multi agent systems (MAS) (VideoAgent \cite{VideoAgent} and videoagent \cite{videoagent2})  and three popular commercial MLLMs (Claude 3.7 Sonnet \cite{anthropic_claude37_sonnet_2025}, GPT-4o \cite{openai_gpt4o_system_card_2024}) and Gemini 2.0 Flash  \cite{google_gemini20_flash_2024}).
For scientific video understanding, \cref{quantitative comparison with sota} shows that SciEducator consistently surpasses all baselines on the relevance and accuracy metrics across the physics, chemistry, and daily‑life tracks, indicating tighter adherence to the visual evidence and underlying scientific principles than general‑purpose MLLMs and prior MASs.

\cref{SciEducator with sota} reports results for the scientific educational content generation task, comparing SciEducator with the closed-source MLLMs (MASs baselines do not support this task). SciEducator demonstrates a clear advantage in creating high-quality instructional materials.

\begin{table}[t]
\centering
\caption{Quantitative Comparison between SciEducator and popular MLLMs and MASs for scientific video understanding on SciVBench. Compared with these SOTA models, our model exhibits higher relevance and accuracy.}
\scriptsize
\setlength{\tabcolsep}{4pt}
\renewcommand{\arraystretch}{1.10}
\setlength{\aboverulesep}{0.2ex}
\setlength{\belowrulesep}{0.2ex}

\begin{tabular}{@{}l*{6}{c}@{}}
\hline
% 表头第一行：Model 使用 multirow 实现上下居中；竖线类型=2（上留白）
\multicolumn{1}{c@{\VSEPtop}}{\multirow{2}{*}{\textbf{Model}}} &
\multicolumn{2}{c}{\textbf{Physics}} &
\multicolumn{2}{c}{\textbf{Chemistry}} &
\multicolumn{2}{c}{\textbf{Daily Life}} \\
\cline{2-7}
% 表头第二行：对应列留空；竖线类型=3（只下留白）
\multicolumn{1}{c@{\VSEPbot}}{} & Rel & Acc & Rel & Acc & Rel & Acc \\
\hline
% —— 分组行：断开竖线 —— 
\multicolumn{1}{c@{}}{} & \multicolumn{6}{c}{Commercial APIs} \\
\hline
% 三行：2 + 4 + 3
\multicolumn{1}{l@{\VSEPtop}}{GPT-4o \cite{openai_gpt4o_system_card_2024}}            & 47.50 & 34.69 & 39.86 & 31.42 & 30.73 & 27.86 \\
\multicolumn{1}{l@{\VSEPnone}}{Gemini 2.0 flash \cite{google_gemini20_flash_2024}}   & 52.81 & 38.75 & 46.96 & 36.15 & 34.64 & 31.25 \\
\multicolumn{1}{l@{\VSEPbot}}{Claude 3.7 Sonnet \cite{anthropic_claude37_sonnet_2025}}   & 44.06 & 31.88 & 40.20 & 31.76 & 31.77 & 28.65 \\
\hline
\multicolumn{1}{c@{}}{} & \multicolumn{6}{c}{Multi-Agent Systems} \\
\hline
% 两行：2 + 3
\multicolumn{1}{l@{\VSEPtop}}{VideoAgent\cite{VideoAgent}}         & 49.06 & 36.56 & 45.61 & 34.80 & 30.47 & 27.34 \\
\multicolumn{1}{l@{\VSEPbot}}{videoagent\cite{videoagent2}}         & 46.25 & 35.31 & 46.62 & 37.16 & 31.51 & 28.13 \\
\hline
% Ours：1（上下都留白）
\multicolumn{1}{l@{\VSEPboth}}{\textbf{SciEducator (Ours)}} &
\textbf{81.88} & \textbf{65.31} & \textbf{73.97} & \textbf{64.86} & \textbf{64.58} & \textbf{62.24} \\
\hline
\end{tabular}
\label{quantitative comparison with sota}
\vspace{-0.2cm} 
\end{table}

% % 若未加载：\usepackage{multirow}
% \begin{table}[t]
% \centering
% \caption{Quantitative Comparison between SciEducator and popular MLLMs/MAS on SciVBench.}
% \scriptsize
% \setlength{\tabcolsep}{4pt}
% \renewcommand{\arraystretch}{1.10}
% \setlength{\aboverulesep}{0.2ex}
% \setlength{\belowrulesep}{0.2ex}

% % 仅保留 Model|Physics 这一条竖线；其它列之间无竖线
% \begin{tabular}{@{}l|*{6}{c}@{}}
% \hline
% \multicolumn{1}{c|}{\multirow{2}{*}{\textbf{Model}}} &
% \multicolumn{2}{c}{\textbf{Physics}} &
% \multicolumn{2}{c}{\textbf{Chemistry}} &
% \multicolumn{2}{c}{\textbf{Daily Life}} \\
% \cline{2-7} % —— Physics 与 Rel/Acc 之间的横线（第2–7列）——
%  & Rel & Acc & Rel & Acc & Rel & Acc \\
% \hline
% % 分组标题只在右侧(第2–7列)中居中；并在该行“断开”竖线
% \multicolumn{1}{c}{ } & \multicolumn{6}{c}{Commercial APIs} \\
% \hline
% GPT-4o            & 46.88 & 32.81 & 38.51 & 30.74 & 29.69 & 27.60 \\
% Gemini 2.0 flash  & 51.25 & 36.88 & 46.96 & 36.15 & 32.81 & 29.43 \\
% Claude 3.7 Sonnet & 43.44 & 28.13 & 38.85 & 30.07 & 30.47 & 27.86 \\
% \hline
% \multicolumn{1}{c}{ } & \multicolumn{6}{c}{Multi-Agent Systems} \\
% \hline
% VideoAgent        & 48.44 & 36.56 & 45.61 & 34.12 & 28.13 & 25.78 \\
% videoagent        & 45.00 & 34.69 & 46.62 & 36.49 & 28.91 & 26.56 \\
% \hline
% \textbf{SciEducator (Ours)} & \textbf{82.50} & \textbf{65.94} & \textbf{74.32} &
% \textbf{65.20} & \textbf{64.58} & \textbf{62.24} \\
% \hline
% \end{tabular}
% \end{table}

\begin{table}[t]
\centering
\caption{Quantitative Comparison between SciEducator and popular MLLMs for scientific education on SciVBench Education Subset. Each metric is reported by win rate (\%).}
\scalebox{0.75}[0.75]{
\begin{tabular}{@{}l|cccc@{}}
\toprule
\multirow{2}{*}{Model} & \multicolumn{4}{c}{Education Subset}       \\ \cmidrule(l){2-5} 
                       & Relevance & IQ    & Attractiveness & EV    \\ \midrule
Gemini 2.0 flash \cite{google_gemini20_flash_2024}       & 10.00     & 2.50  & 0.00           & 5.00  \\
GPT-4o \cite{openai_gpt4o_system_card_2024}               & 7.50      & 5.00  & 2.50           & 7.50  \\
Claude 3.7 Sonnet \cite{anthropic_claude37_sonnet_2025}     & 5.00      & 5.00  & 0.00           & 5.00  \\
SciEducator(Ours)      & \textbf{77.50}     & \textbf{87.50} & \textbf{97.50}          & \textbf{82.50} \\ \bottomrule
\end{tabular}}
\label{SciEducator with sota}
\vspace{-0.5cm} 
\end{table}

\paragraph{Qualitative Analysis.}
As illustrated in \cref{Qualitative with sota video understanding}, for specific user queries, SciEducator provides responses that are more comprehensive, detailed, and clearly explained compared to MLLMs such as GPT-4o \cite{openai_gpt4o_system_card_2024}. This advantage stems from two core capabilities of SciEducator: first, the integration of multiple powerful search tools significantly expands the system's knowledge capacity; second, its robust planning and replanning capabilities enable it to better analyze complex video content and acquired information, resulting in more organized and higher-quality answers.

As shown in \cref{fig:Figure 1}, SciEducator can generate e-learning materials incorporating four modalities: text, images, hyperlinks, and audio, featuring rich colors and aesthetically pleasing layouts. The text is structured with sequential sections including a title, introduction, experimental procedures and precautions, and a summary. As illustrated in \cref{Qualitative with sota education}, this structural arrangement effectively guides readers through understanding the principles and completing the experiment. The visual elements include images of required materials and model-generated diagrams illustrating experimental steps, providing clearer guidance than text alone and highlighting SciEducator's significant advantage over conventional LLMs.

% \subsubsection{Efficiency Analysis}
% \textcolor{red}{SciEducator averages [t] s/item, [m] tool calls,campared to VideoAgent's  [t1] s,and commercial MLLM [t2] s.}

\subsection{Ablation Studies}

\begin{figure}[t]
    \centering
    \includegraphics[width=1.0\linewidth]{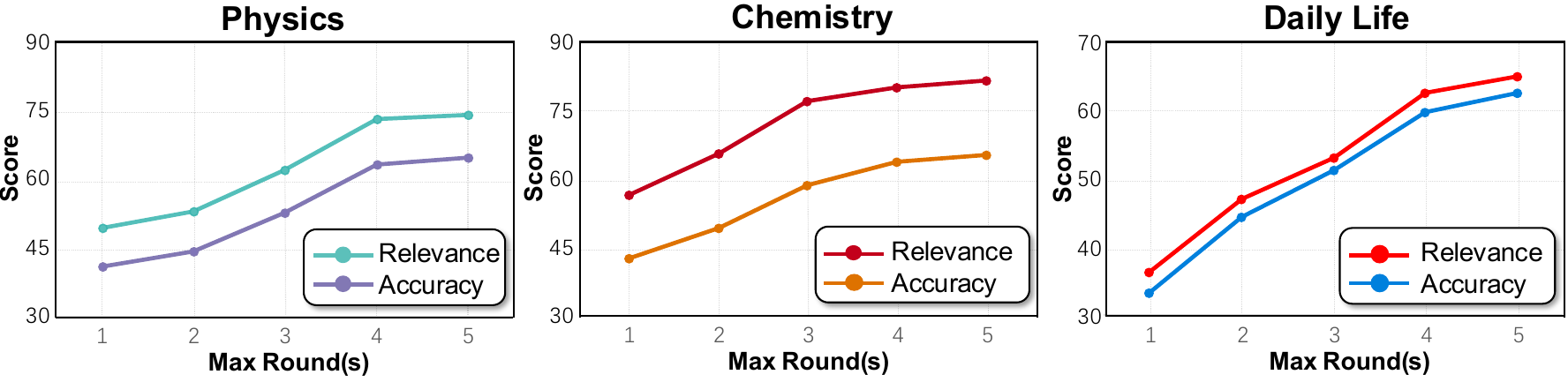}
    \caption{Ablations on different maximum iteration rounds for scientific video understanding. Performance rises with larger Max Rounds for three categories, visualizing the benefit of iterative PDSA cycles. }
    \label{ablation_study_max_round}
    \vspace{-0.2cm} 
\end{figure}

\begin{table}[t]
\centering
\caption{Ablations on different maximum iteration rounds for education. Each metric is reported by win rate (\%).}
 \scalebox{0.73}[0.77]{
\begin{tabular}{@{}l|c|cccc@{}}
\toprule
\multirow{2}{*}{Model} & \multirow{2}{*}{Max Round(s)} & \multicolumn{4}{c}{Education Subset}       \\ \cmidrule(l){3-6} 
                       &                               & Relevance & IQ    & Attractiveness & EV    \\ \midrule
SciEducator            & 1                             & 2.50      & 0     & 2.50           & 15.00 \\
SciEducator            & 3                             & 7.50      & 7.50  & 32.50          & 35.00 \\
SciEducator            & 5                             & \textbf{90.00}     & \textbf{92.50} & \textbf{65.00}          & \textbf{50.00} \\ \bottomrule
\end{tabular}}
\label{SciEducator with Different Maximum Iteration Rounds}
\vspace{-0.2cm} 
\end{table}

\paragraph{Effectiveness of PDSA Cycles.}
To validate the progressive improvement of SciEducator in response relevance and accuracy through PDSA cycles, we configured the maximum number of cycles from 1 to 5. After reaching the cycle limit, the system was compelled to generate the most likely correct answer based on accumulated knowledge. Results in \cref{ablation_study_max_round} show that performance improves significantly with more cycles, demonstrating SciEducator’s capacity for self-evolution and its ability to iteratively deepen understanding, acquire relevant knowledge, and solve unconventional, complex problems.

To validate the progressive improvement of SciEducator in generating educational text via PDSA cycles, we compare three variants with maximum cycle counts of 1, 3, and 5 on the Education Subset of SciVBench, with results also presented as win rates across various evaluation aspects and summarized in \cref{SciEducator with Different Maximum Iteration Rounds}. Results demonstrate that as the PDSA cycles increase, the model's performance improves, yielding instructional text with stronger relevance and more educational value, which in turn leads to the generation of higher-quality guidance images.

\paragraph{Effectiveness of Evaluator Agent.}

To validate that each evaluation element in the Evaluator Agent effectively reduces resource consumption and execution rounds, we compared the complete EA with ablated versions lacking these elements on 500 QA pairs. As shown in \cref{ablation study with EA}, metrics include average time, token consumption, number of execution rounds, and final accuracy, with consumption values shown as ratios. Note that the total consumption of the complete EA is normalized to 1.00. The maximum number of execution rounds was set to 5. Experimental results demonstrate that each evaluation element in the Evaluator Agent effectively increases the probability of selecting superior solutions, reduces resource consumption and identifies more feasible options, and consequently generates responses with higher accuracy within limited iteration cycles.

\paragraph{Effectiveness of $K_{\text{new}}$ \& $F$.}
To validate the effectiveness of new knowledge $K_{\text{new}}$ and failure analysis $F$ of the Study Stage, which updates the solution pool and improves the quality of candidate solutions,  we conducted ablations by removing these elements, as presented in \cref{ablation study with Study Stage}. In the variant where both $K_{\text{new}}$ and $F$  are removed, no feedback is available, and updates only discard executed solutions. The experimental results indicate that $K_{\text{new}}$ and $F$ are critical for optimizing existing solutions and generating new ones. Within the PDSA process, these elements substantially improve the quality of the new solution pool, thereby enhancing both the relevance and accuracy of the results.

% \textcolor{red}{addition experiments I think need to add}

% \textcolor{red}{1. 6 or 7 cycles resulted in only a minor improvement in effect, but consumed much more time resources. Considering the trade-off between effect and efficiency, we ultimately chose 5 cycles.}

% \textcolor{red}{2.More visualizations of the ablation results, such as what is the output of the first cycle, what is it for the third cycle, and what is it for the fifth cycle.}

% \textcolor{red}{3. Conduct ablation experiments to remove the effects of some minor modules, such as these 9 dynamically acting agents and 7 fixed tool or agents.
% }

% \textcolor{red}{4. Conduct ablation experiments to remove the effects of RAG and remove the effects of web search
% }

% \textcolor{red}{5. Conduct ablation experiments about the evaluator, (1).no IDF ,just $\mathcal{E}$, (2) no $\mathcal{E}$,just IDF, (3)$\mathcal{E}$ and IDF 
% }

\begin{table}[t]
\centering
\caption{Ablations on Evaluator Agent (EA). Note that $\mathcal{E}$ denotes empirical knowledge and $A_{\text{percep}}$ represents Perceptual Evaluation metrics. Time and token consumption values are presented as ratios, with the total consumption of the complete EA version normalized to $1.00$.}
\scalebox{0.82}[0.75]{
\begin{tabular}{@{}l|ccc|c@{}}
\toprule
Agent               & \ Time$\downarrow$ & 
\ Token$\downarrow$ & Average Rounds$\downarrow$ & \ Acc $\uparrow$  \\ \midrule
EA w/o $\mathcal{E}$ & 1.20 & 1.18  & 4.09           & 57.50 \\
EA w/o IDF          & 1.08 & 1.06  & 3.99           & 59.90 \\
EA w/o $A_{\text{percep}}$           & 1.14 & 1.13  & 4.17           & 54.50 \\ \midrule
EA                  & \textbf{1.00} & \textbf{1.00}  & \textbf{3.79}           & \textbf{64.00} \\ \bottomrule
\end{tabular}}
\label{ablation study with EA}
\vspace{-0.1cm} 
\end{table}

\begin{table}[t]
\centering
\caption{Ablations of on the effectiveness of different elements in the Study Stage. $K_{\text{new}}$ represents the new knowledge acquired during execution, and $F$ denotes failure analysis.}
\scalebox{0.70}[0.70]{
\begin{tabular}{@{}l|cccccc@{}}
\toprule
\multirow{2}{*}{Model}             & \multicolumn{2}{c}{Physics} & \multicolumn{2}{c}{Chemistry} & \multicolumn{2}{c}{Daily Life} \\ \cmidrule(l){2-7} 
                                   & Rel          & Acc          & Rel           & Acc           & Rel            & Acc           \\ \midrule
SciEducator w/o $K_{\text{new}}$ \& $F$ & 59.69        & 45.94        & 53.04         & 45.27         & 35.94          & 32.55         \\
SciEducator w/o $K_{\text{new}}$          & 65.94        & 50.94        & 61.82         & 54.05         & 38.28          & 34.64         \\
SciEducator w/o $F$              & 71.56        & 55.63        & 66.55         & 57.09         & 48.95          & 45.83         \\ \midrule
SciEducator                        &\textbf{81.88}        & \textbf{65.31}     & \textbf{73.97}       &\textbf{64.86}    & \textbf{64.58}          & \textbf{62.24}      \\ \bottomrule
\end{tabular}}
\label{ablation study with Study Stage}
\vspace{-0.5cm} 
\end{table}

\section{Conclusion}
In this work, we present \name{}, the first iterative self-evolving multi-agent system for scientific video comprehension and education. By reformulating the Deming Cycle’s \emph{Plan–Do–Study–Act} philosophy into a self-evolving reasoning and feedback mechanism, the system progressively enhances its interpretation of complex scientific activities to deliver accurate and reliable results. To ensure comprehensive performance assessment, we construct the \dataset{} benchmark. Extensive experiments show that \name{} significantly outperforms current popular models, demonstrating remarkable superiority and application potential, and is expected to offer meaningful inspiration for future multi-agent systems and broader domains.

%% file: sec/3_finalcopy.tex
% \section{Final copy}

% You must include your signed IEEE copyright release form when you submit your finished paper.
% We MUST have this form before your paper can be published in the proceedings.

% Please direct any questions to the production editor in charge of these proceedings at the IEEE Computer Society Press:
% \url{https://www.computer.org/about/contact}.

%% file: sec/X_suppl.tex
\clearpage
\setcounter{page}{1}
\renewcommand{\thesection}{\Alph{section}}
\setcounter{section}{0} % 从A开始计数
\maketitlesupplementary

\section{Tools \& Agents}
Our system integrates 10 agents and 6 tools, each designed to handle specific tasks with clearly defined input and output specifications to ensure precise execution and seamless integration. The set of tools and agents is categorized into two groups: (i) dynamically invocable components and (ii) fixed-execution components, described as follows.
\subsection{Dynamically Invocable Tools/Agents}
\textbf{(i) Planning and Answer Generation (Planner Agent)}: Our system uses a Planner Agent configured with GPT-4o as its core. Leveraging the powerful text processing and knowledge reasoning capabilities of the LLM, it is responsible for generating and adjusting the solution pool within the system, consolidating all acquired information, evaluating confidence level, and outputting the final answer. 

\textbf{(ii) Video Content Acquisition (Captioner Agent)}: Our system uses a Captioner Agent configured with Gemini 2.0 Flash to obtain the content of video frames and generate textual descriptions of the video.

\textbf{(iii) Solution Evaluation (Evaluator Agent)}: Our system uses an Evaluator Agent configured with GPT-4o to evaluate the various solutions in the solution pool. It selects the best solution from the current pool by considering both subjective and objective metrics and initiates its execution. 

\textbf{(iv) Web Content Search (Web Search Agent)}: We use the open-source web content search model SegGPT, based on the Google Search engine, to search for web links related to input keywords and organize the web information for output. 

\textbf{(v) Paper Search (Paper Search Agent)}: We develop a Paper Search Agent configured with GPT-4o for searching academic paper content. It can search for papers related to input keywords across major scientific paper platforms, organize and summarize the found content, and finally output the results. 

\textbf{(vi) Video Super-Resolution Tool (VideoSR Tool)}: We built a tool for performing super-resolution on video frames based on an open-source model. This enhances low-resolution or blurry frames, ultimately improving the description of the video content. 

\textbf{(vii) Experimental Procedure Search (Procedure Search Agent)}: We develop a Procedure Search Agent based on the open-source web content search model searchGPT. It searches for corresponding experimental procedures based on input experimental terms or descriptions, organizes the web information, and outputs the results.

\textbf{(viii) Key Entity Recognition in Content (Entity Recognition Agent)}: SciEducator uses an Entity Recognition Agent configured with GPT-4o to identify key experimental instruments and materials involved in the current experimental steps, and outputs them in a specified format. 

\textbf{(ix) Experimental Precautions Prompter (Safety Alert Agent)}: We develop a Safety Alert Agent based on the open-source web content search model searchGPT. It comprehensively searches for corresponding safety precautions based on the current experimental content and equipment, organizes the information, and outputs it to alert readers to avoid potential hazards when conducting the experiment.

\subsection{Fixed-Execution Tools/Agents}
\textbf{(i) Knowledge Base Construction and Storage}: We develop tools for knowledge base construction and storage. These tools archive knowledge texts, invoke an embedding model to create and store query vectors, thereby facilitating subsequent retrieval. The construction and storage of the knowledge base must be completed before the DCAgent operates. 

\textbf{(ii) Knowledge Base Retrieval (RAG Agent)}: We construct a small-scale knowledge base containing basic scientific knowledge and fundamental material properties (e.g., electromagnetic induction, properties of air). We configured a Retrieval-Augmented Generation (RAG) Agent based on GPT-4o. Based on the video content, it can extract important keywords, retrieve the most relevant content from the knowledge base based on these keywords, and output an organized summary. If no relevant content exists in the knowledge base, it will output a statement indicating the lack of relevant knowledge. 

\textbf{(iii) IDF Value Calculation for Keywords in Solutions (IDF Calculator Tool)}: We independently develop a tool capable of calculating the Inverse Document Frequency of each keyword within the knowledge base, indicating the uniqueness of each keyword and its importance to the video content. It is fixedly invoked each time the Evaluator Agent evaluates the solutions in the solution pool. The Evaluator Agent extracts keywords from each solution and inputs them into the IDF Calculator Tool for computation.

\textbf{(iv) Experimental Equipment Image and Purchase Link Search (Equipment Search Tool)}: We independently develop a tool for searching images and purchase links of experimental equipment, typically returning one image and one most relevant link per equipment item. 

\textbf{(v) Experimental Procedure Illustration Generation (Illustration Generation Tool)}: We utilize the Gemini 2.5 Flash Image (Nano Banana) API to generate illustrations of experimental procedures by inputting descriptive text of the steps. 

\textbf{(vi) Text-to-Speech Conversion (Speech Generation Tool)}: We employ the open-source tool KittenTTS to convert experiment-related text into speech. 

\textbf{(vii) Multi-modal Information Integration for Structurally Organized and Aesthetically Arranged E-booklet Compilation (E-booklet Generation Agent)}: We independently developed an E-booklet Generation Agent for integrating multi-modal information related to experiments, such as text, images, links, and audio. Implemented in JavaScript/CSS,the agent composes an HTML e-booklet with clear hierarchy and an aesthetically pleasing layout to stimulate readers' scientific interest in a vivid and engaging manner.

\section{Details about Empirical Prior}

To obtain the resource consumption and the probability of returning expected results for dynamically invocable tools/agents, we build an empirical prior $\mathcal{E}$ by issuing 20 randomized probe calls per tool/agent, from which Evaluator Agent obtains average latency, average token usage, and success probability.  Token usage is directly converted into the corresponding platform’s API money cost. Results: 

Web Search Agent: On average, each call takes 21.89s, costs \$0.0139, and 88\% success rate.

Paper Search Agent: On average, each call takes 17.41s, costs \$0.010, and  75\% success rate.

Captioner Agent: On average, each call takes about 25s if the frame rate is 1 FPS, costing \$0.0001; both time and financial cost increase in direct proportion to the FPS, 93\% success rate.

VideoSR Tool: Approximately 53s if the frame rate is 1 FPS, costing \$0.0001; both time and financial cost increase in direct proportion to the FPS, 100\% success rate.

Procedure Search Agent: On average, each call takes 28.17 seconds, costs \$0.0178, and  87\% success rate.

Entity Recognition Agent: On average, each call takes 10.45 seconds, costs \$0.0064, and  99\% success rate.

Safety Alert Agent: On average, each call takes 26.29 seconds, costs \$0.0153, and 72\% success rate.

Based on our practical considerations, we combine time consumption (t) and financial cost (c) into a total cost: t + 1000c. This is provided to the Evaluator Agent as a reference. The Evaluator Agent is instructed to give equal weight to resource consumption and feasibility considerations, and ultimately to select the overall optimal solution.

\section{Evaluation Metrics}
\subsection{Understanding}

We employ Qwen3-Max to uniformly evaluate all model-generated responses. We first provide the reference answers and substantial scientific background regarding the query questions, instructing Qwen3-Max to assess the relevance of the model-generated answers to the query, focusing on how well the response aligns with the scientific subdomain involved in the question, regardless of correctness. The objective is to determine whether the model provides misleading or irrelevant information. The detailed scoring strategy is as follows:

\begin{itemize}
\item 1.If the answer is entirely relevant to the subfield of the question, it receives a relevance score of 1.
\item 2.If an answer contains partially irrelevant content or is only broadly related to the field, it receives a score of 0.5.
\item 3.If it is completely irrelevant, it receives a score of 0.
\end{itemize}

\begin{notebox}[breakable,enhanced,title=Prompt: Evaluating Relevance]

        You are a professional scoring teacher and evaluation expert. Your task is to evaluate the relevance of responses based on the reference answer and the scientific background related of question, using three scoring options: 1 point, 0.5 points, or 0 points.

        \vspace{\baselineskip} % 一个行高的间距

        Please follow these scoring criteria in order of priority from highest to lowest:

          \vspace{\baselineskip} % 一个行高的间距
        1.If the answer is entirely relevant to the specific subfield of the question, award 1 point.
        
        2.If the answer contains partially irrelevant content or is only broadly related to the field of the question, assign 0.5 points.
        
        3.If the answer is completely irrelevant, assign 0 points.

        \vspace{\baselineskip} % 一个行高的间距
        Please provide the points directly, only the number, no other output.

\end{notebox}

We then use Qwen3-Max to analyze the semantic similarity between generated answers and reference answers, scoring the accuracy of model responses. The detailed scoring strategy is as follows:
\begin{itemize}
\item 1.If the model's response contains absolute errors—including numerical inaccuracies (e.g., the correct answer is 3, but the model answers 2) or terminological mistakes (e.g., the correct answer is "Magnus Effect," but the model answers "Bernoulli’s Principle")—it receives a score of 0 directly.
\item 2. Qwen3-Max is prompted to analyze the key points of the reference answer and determine whether the model's response covers all of them. If no key point is covered, the score remains 0. If some key points are covered and the remaining content contains no absolute errors, a score of 0.5 is assigned.
\item  3. If all key points are covered and no absolute errors are present in the remaining content, a full score of 1 is awarded.
\end{itemize}

\begin{notebox}[breakable,enhanced,title=Prompt: Evaluating Accuracy]

        You are a professional scoring teacher and evaluation expert. Your task is to evaluate the quality of responses based on reference answers, using three scoring options: 1 point, 0.5 points, or 0 points.
        \vspace{\baselineskip} % 一个行高的间距
        
        Please follow these scoring criteria in order of priority from highest to lowest:

          \vspace{\baselineskip} % 一个行高的间距
        1. First check for any absolute errors that contradict the reference answers, such as instances where the reference answer deems something correct but the response considers it incorrect, or numerical mistakes; if such errors occur, award 0 points.

        2. Then analyze the key points of the reference answers and compare them to the responses, noting that wording can differ but the meaning must be strictly identical to count as a match; if the response fully covers all key points of the reference answers, give 1 point, and additional correct content without errors can still warrant 1 point.

        3. Finally, if the response matches only part of the key points and contains no absolute errors, assign 0.5 points, but exercise caution when awarding this score. Don't be too rigid, you should fully refer to the different expressions of the standard answer. If there are more details than the standard answer, it should be considered correct
        
        \vspace{\baselineskip} % 一个行高的间距
        Please provide the points directly, only the number, no other output.

\end{notebox}

\subsection{Educating}

We employ four metrics and uniformly use Qwen-VL-Plus to evaluate all model responses in a comparative setting. Qwen-VL-Plus receives supplied with substantial background information about each experiment as a reference. Specifically, the metrics are:
\begin{itemize}
    \item \textbf{Relevance}: How well the generated experimental procedures and precautions align with the current experiment and its underlying principles.
    \item \textbf{Instructional Quality (IQ)}: How effectively the generated procedures and precautions guide children in conducting the experiment, with emphasis on detail orientation, completeness, clarity, and safety warnings.
    \item \textbf{Attractiveness}: A comprehensive assessment of how engaging the textual instructions are. For SciEducator, the aesthetic quality of its supporting illustrations is also incorporated into this evaluation to identify the most captivating response.
    \item \textbf{Educational Value (EV)}: How well each model’s response stimulates children’s scientific interest and guides them to understand the principles through the experiment.
\end{itemize}

\begin{notebox}[breakable,enhanced,title=Prompt: Evaluating Performance in Education]

You are a fair and professional evaluation expert. Several models have generated experimental procedures and safety precautions for a specific scientific phenomenon. Your task is to compare the responses produced by these different models from four aspects and select the best-performing model for each aspect.

    \vspace{\baselineskip} % 一个行高的间距
You will be provided with a description of the scientific phenomenon and substantial background information about it as a reference. The four aspects are as follows:

    \vspace{\baselineskip} % 一个行高的间距
Relevance: How well each model’s generated experimental procedures and precautions align with the current experiment and its underlying principles.

Instructional Quality: How effectively each model’s generated procedures and precautions guide children in conducting the experiment, with emphasis on detail orientation, completeness, clarity, and safety warnings.

Attractiveness: How engaging the responses are. If any response contains images, the images should also be included in the evaluation.

Educational Value: How well each model’s response stimulates children’s scientific interest and guides them to understand the principles through the experiment.

\vspace{\baselineskip} % 一个行高的间距
Please select the best-performing model for each aspect, referring to the models by their names, and output the result in the following format:

[

     \{

        "Relevance": model name,
        
        "Instructional Quality": model name,
        
        "Attractiveness": model name,
        
        "Educational Value": model name

    \}

    ]
\end{notebox}

\section{Main Prompts}
\subsection{Plan Stage}
\begin{examplebox}[breakable,enhanced,title=Prompt: Plan Stage]

"""

    User Query: \{user\_query\}
    
    Video Path: \{video\_path\}
    
    Video Content Description: \{video\_description\}
    
    RAG Search Results: \{rag\_result\}

    \vspace{\baselineskip} % 一个行高的间距
    You are a scientific video understanding task planning expert. Your responsibilities are:
    
    1. Generate multiple possible solutions based on video information, user queries, and retrieved knowledge
    
    2. Each solution should include specific descriptions, tool call sequences, and parameters for each tool input

    \vspace{\baselineskip} % 一个行高的间距
    Please output a list of solutions directly in JSON format without any additional text:

    [
    
         \ \ \{
        
            "Number": Solution number (integer),
            
            "description": "Solution description (detailed explanation of the solution process)",
            
            "steps": [
            
                \ \ \ \ \ \ \ \ \{
                
                "tool": "Name of the tool to call (must use one of the following recognized tool names: WebSearch, PaperSearch, Captioner, VideoSR)",
                
                "input": "Tool input parameters (clear and specific input content)"
                
                \ \ \ \ \ \ \ \ \}

           \ \ \ \ \ \ \ \ \ \ \  ]
            
       \ \ \}
        
    ]

   \vspace{\baselineskip} % 一个行高的间距
    Tool Description:
    
    - WebSearch: Web search tool, can input a query sentence
    
    - PaperSearch: Academic paper search tool, can input a query sentence

    - VideoProcessor: Video understanding tool, inputs include a video path, number of segments to divide the video into, frames per second to process, and information to search for in the video
    
    - VideoSR: Video super-resolution tool, input parameters are the same as VideoProcessor, performs super-resolution on input frames before understanding

    \vspace{\baselineskip} % 一个行高的间距

    Do not include any other fields or text, output only JSON objects.

    \vspace{\baselineskip} % 一个行高的间距
    Note:
    
     1. Your solutions should be as logical and reasonably sequenced as possible, be relevant to user queries, striving to preserve the substances, environment, and specific movements of phenomena occurring in the video. Including the experimental objects and environment in tool input parameters may increase the success probability of the solution. 
     
     2. You should output as many possible solutions as possible to facilitate unified evaluation and improve success rate. Your output solutions may have more details, be more complete, and have longer call chains than our examples.
     
     3. Please carefully check whether each step of your solution is conducive to solving the problem, such as querying the miraculous movements, changes, or interactions of objects, the environment or solution in which phenomena occur, etc. Make good use of each tool to obtain information and reduce ineffective calls.
     
     4. Do not rashly conclude what phenomenon the video describes before you have obtained sufficient information. The parameters input to all tools in your solution should be as specific as possible. For example, if the video phenomenon occurs in mate tea, and you want to query the interaction between solid and liquid, or the peculiar movement of solids in liquid, we recommend that you always change the liquid here to mate tea, i.e., "peculiar movement of solids in mate tea".
     
     5. Note that you are working on a scientific phenomenon understanding task. Your solutions should aim to clarify the scientific phenomena in the video content description.

     \vspace{\baselineskip} % 一个行高的间距
     Please strictly adhere to the requirements
     
    """
\end{examplebox}

\subsection{Do Stage}
\begin{examplebox}[breakable,enhanced,title=Prompt: Do Stage]

"""

 You are a professional solution evaluation expert. Please evaluate multiple solutions and select the best one based on the following information:
 
    User Query: \{user\_query\}
    
    RAG Search Results: \{rag\_result\}
    
    Previous Solution Results: \{previous\_results\}

    Here are the solutions to evaluate along with their keyword IDF values (higher IDF values indicate more unique keywords):
    \{plans\_with\_idf\}

    \vspace{\baselineskip} % 一个行高的间距
    Please follow these evaluation steps:
    1. Analyze the feasibility and relevance of each solution. Evaluate feasibility and relevance according to the following criteria:
    
    2. Consider the IDF values of keywords - higher IDF keywords may be more representative. Then consider the frequency of keywords in the video content (TF values) - higher TF keywords may be more representative. 
    
    3. Evaluate the completeness and logical coherence of each solution. Evaluate whether each solution  contains scientific phenomena — such as "Why can a person swing higher despite no energy being added?"
    
    4. Comprehensively consider the time consumption and financial cost of tools called in each solution, along with the probability of returning expected results of them. Below is the tool information list:
    
       WebSearch: 21.89s, \$0.0139, 88\% success rate
       
       PaperSearch: 17.41s, \$0.010, 75\% success rate
       
       VideoProcessor: 25s if the frame rate is 1 FPS, costing \$0.0001; both time and financial cost increase in direct proportion to the FPS, with a 93\% success rate.
       
       VideoSR: 53s if frame per second = 1, \$0.0001, both time and financial cost increase in direct proportion to the FPS, with a 100\% success rate.

        \vspace{\baselineskip} % 一个行高的间距
       We roughly combine time consumption (t) and financial cost (c) into total consumption: t + 1000c. You can reference this value for judgment

     \vspace{\baselineskip} % 一个行高的间距   
    5. Considering all steps, giving equal weight to resource consumption and feasibility considerations, and ultimately to select the overall optimal solution. 

     \vspace{\baselineskip} % 一个行高的间距
    Please output the best solution directly in JSON format without any additional text:

    \{
    
        "best\_plan": \{
        
            "Number": "solution number",
            
            "description": "solution description",
            
            "Why": reasons you choose it as the best plan and reasons other plans are not good
            
            "steps": [
            
              \ \ \ \ \ \ \  \{
              
                    "tool": "tool name",
                    
                    "input": "input parameters"
                    
                \ \ \ \ \ \ \  \}
                
            ]
            
        \ \ \ \ \ \ \  \ \ \ \ \ \ \ \ \ \ \ \ \ \ \}
        
    \}

    """
\end{examplebox}

\subsection{Study \& Act Stage}
\begin{examplebox}[breakable,enhanced,title=Prompt: Study \& Act Stage]

"""

    User Query: \{user\_query\}
 
    Video Path: \{video\_path\}
    
    Video Content: \{video\_descriptions\}

    RAG Search Results: \{rag\_result\}
    
    Historical Execution Results and Current Execution Result: 
    \{history.get('previous\_results', [])\}
    
    All Available Solutions: \{json.dumps(all\_plans, ensure\_ascii=False, indent=2)\}
    
    executing plan number:\{Number\}

       \vspace{\baselineskip} % 一个行高的间距
    Please analyze the reasons for the failure or poor outcome of this execution, summarize the existing execution results and new knowledge/information, adjust previous solutions, and attempt to generate completely new solutions. Finally, compile them into a new solution collection called new\_plans.
    
   \vspace{\baselineskip} % 一个行高的间距
    Return a JSON object containing:
    
    \{
    
        "failure\_analysis": "Analysis of failure reasons",

        "knowledge\_summary": "Summary of knowledge and information contained in existing execution results",

        "new\_plans": "List of new solutions (same format as the original solution list)"
        
    \}

    \vspace{\baselineskip} % 一个行高的间距
    Tool Description:
    
    - WebSearch: Web search tool, can input a query sentence
    
    - PaperSearch: Academic paper search tool, can input a query sentence
    
    - VideoProcessor: Video understanding tool, inputs include a video path, number of segments to divide the video into, frames per second to process, and information to search for in the video
    
    - VideoSR: Video super-resolution tool, input parameters are the same as VideoProcessor, performs super-resolution on input frames before understanding

    \vspace{\baselineskip} % 一个行高的间距
    Here are some examples you can reference for adjusting or generating solutions based on failure reasons:
    
    1.If you find that video information is insufficient, you can adjust or generate solutions by calling the VideoProcessor tool, increasing the input frame rate, and reducing the number of segments to obtain more information. However, you should not let VideoProcessor continue searching for things in existing solution as you might be going off track
    
    2.If you need specific video information to determine the final answer, you can adjust or generate solutions by calling the VideoProcessor tool and specifying the questions you want to query
    
    3.If the VideoProcessor tool returns that the video is blurry and affecting answer generation, you can call VideoSR to clarify the video and then understand.
    
    4.If WebSearch or PaperSearch did not return expected information or timeout, you can try adjusting your search keywords
    
    5.If the search returns information that is too broad, you can try using more precise keywords for searching
    
    6.If the search returns several completely different scenarios and you need more details to determine the final answer, you can call appropriate tools to obtain them
    
    7.If a particular tool suddenly fails, you should reduce the use of that tool

     \vspace{\baselineskip} % 一个行高的间距
    Note: You don't necessarily have to generate new solutions every time; it's also acceptable to simply discard failed solutions.
    
    """
\end{examplebox}

\section{More Visualization Result}
In this section, we provide additional visualization results to further demonstrate SciEducator's capabilities in both scientific video understanding and educational content generation.
\subsection{Scientific Video Understanding}
\begin{figure*}[t]
    \centering
    \includegraphics[width=1.0\linewidth]{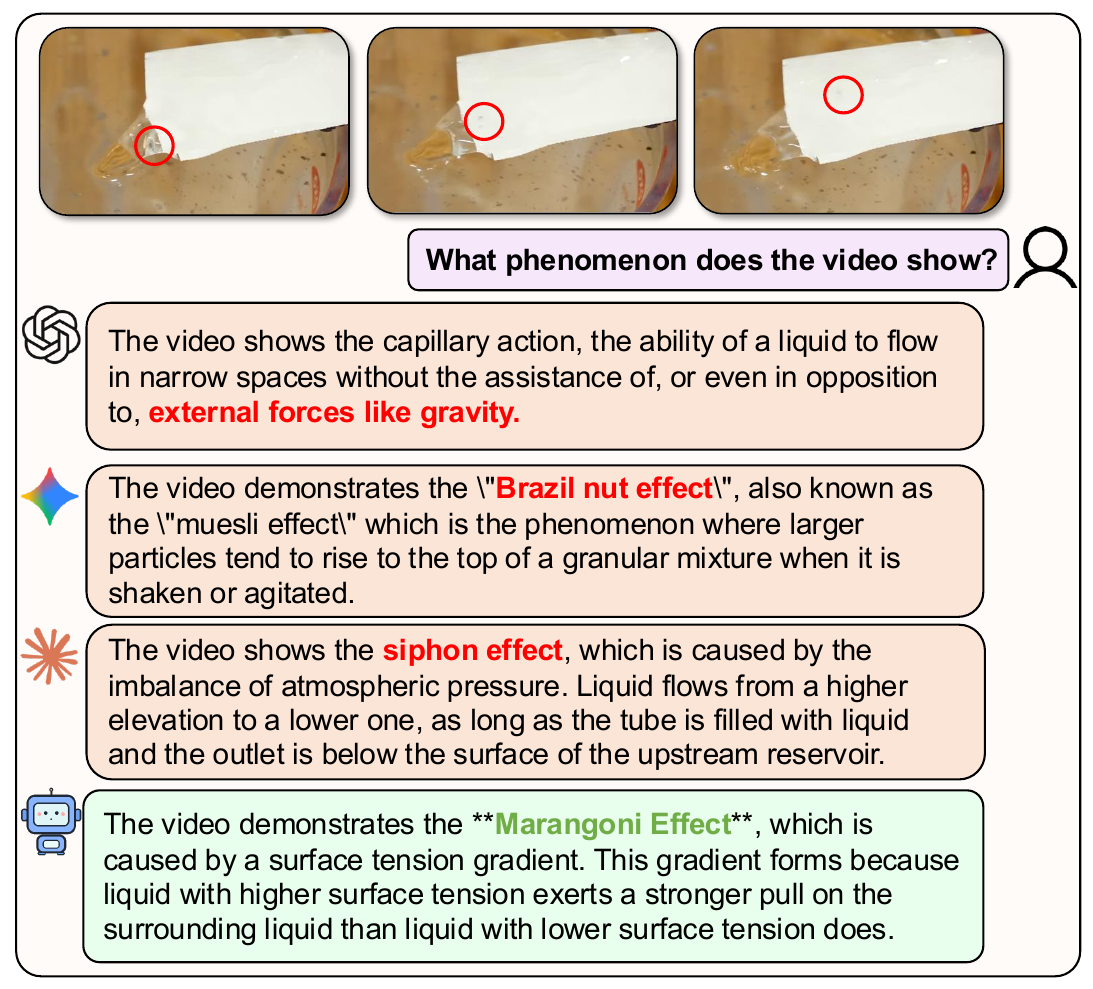}
    \caption{Qualitative comparison between SciEducator and MLLMs. These examples demonstrate SciEducator's ability to generate more comprehensive, better-structured, and more logically coherent answers than the other MLLMs. }
    \label{Qualitative with sota video understanding supple}
\end{figure*}

SciEducator leverages a unique iterative self-evolving mechanism rooted in the Deming Cycle (Plan-Do-Study-Act) to achieve rigorous step-wise reasoning \cref{Qualitative with sota video understanding supple}. Unlike standard MLLMs that may hallucinate or provide superficial answers, our system integrates external professional knowledge and performs failure analysis to refine its understanding. This capability serves as the foundation for the subsequent educational stage, ensuring that the scientific principles identified (e.g., light refraction, chemical reactions) are accurate before being translated into educational materials.

\subsection{Educational E-booklet Generation}
A core innovation of SciEducator is its ability to generate comprehensive, child-friendly educational E-booklets, which is shown in \cref{fig:booklet-a}. The system organizes multimodal content—including text, diagrams, and safety alerts—into a structured format that fosters engagement and safety. The full E-booklet is visualized in five segments:

\begin{itemize}
\item The booklet begins with an engaging title and an "Interesting Introduction" designed to capture the learner's curiosity. As shown in the visualization, the system uses evocative language (e.g., "a tiny scientist, exploring the wonders") to transform complex scientific concepts into an accessible narrative, setting the stage for the experiment.

\item The E-booklet describes the detailed experimental materials and corresponding pictures and shopping links, which are helpful for the rapid commencement of the experiment.

\item The  E-booklet provides a step-by-step, detailed description of the specific experimental procedures, along with corresponding illustrations, to assist in conducting the experiments.
\item The  E-booklet also provides important notes, which highlight some dangerous actions and operations that may lead to experimental failures, thereby ensuring the safety of experiments.
\item The E‑booklet concludes with a concise summary that reinforces core concepts and takeaways.
\end{itemize}

\clearpage % 强制换页，确保从新的一页开始
\begin{figure*}[htbp] 
    \centering
    \includegraphics[width=0.95\linewidth]{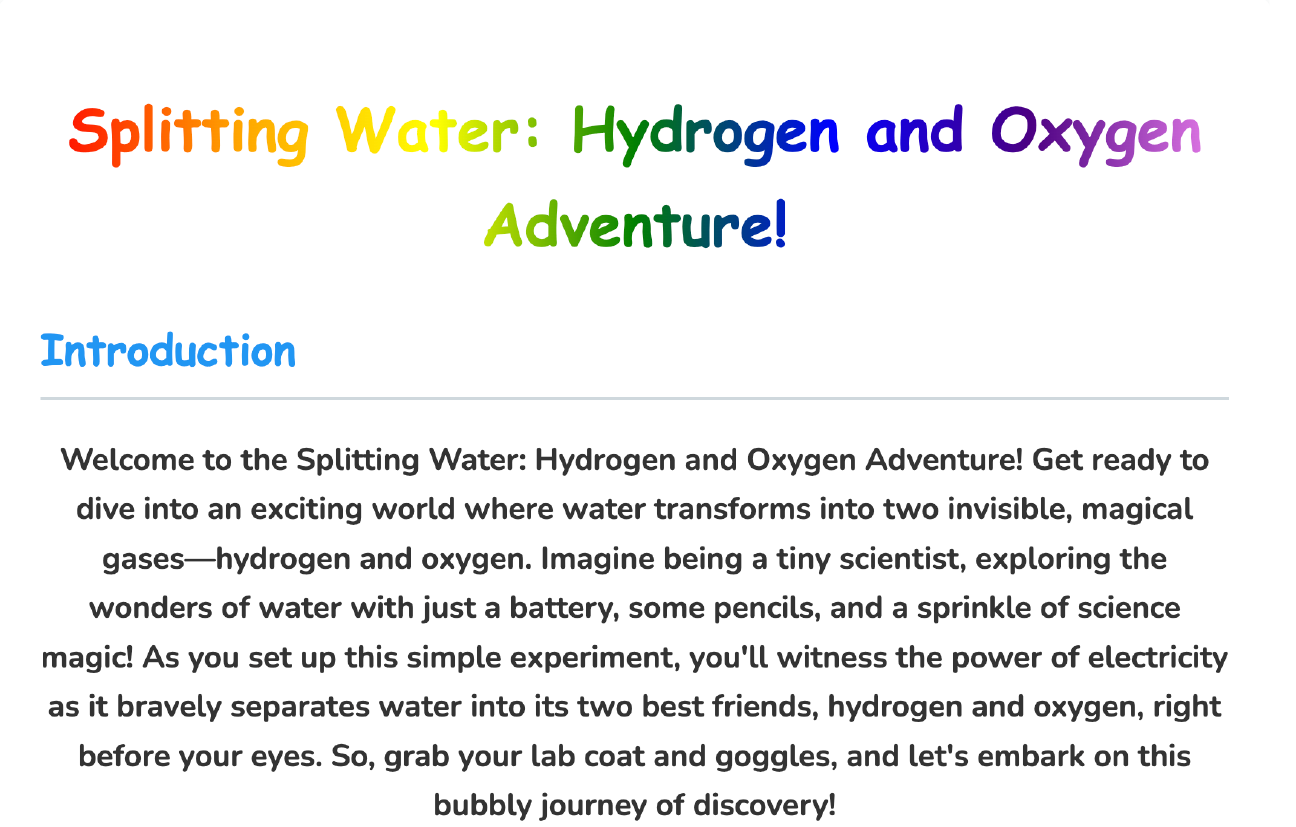} % 0.95留一点边距，防止溢出
    \caption{Our generated e-booklet with comprehensive contents and a well-organized structure. These examples demonstrate SciEducator's ability to generate more comprehensive, better-structured, and more attractive Education E-booklet. \\ \textbf{(a) About Title and Interesting Introduction}} 
    
    \label{fig:booklet-a}
\end{figure*}

\clearpage
\begin{figure*}[htbp]
    \ContinuedFloat % 延续上一张图的编号
    \centering
    \includegraphics[width=0.8\linewidth,height=0.85\textheight]{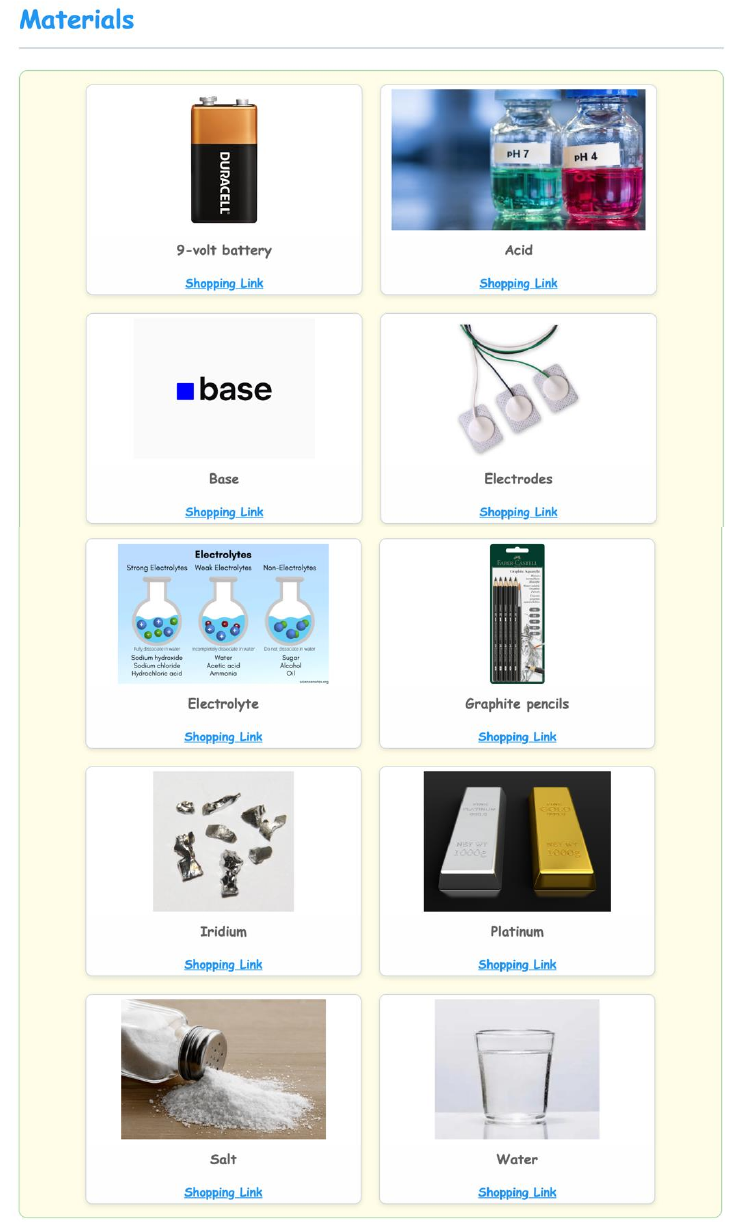}
    \caption[]{Our generated e-booklet (Continued). \\ \textbf{(b) Experiments Materials List}}
    \label{fig:booklet-b}
\end{figure*}

\clearpage
\begin{figure*}[htbp]
    \ContinuedFloat
    \centering
    % 已自动修正文件名，补上了 .pdf
    \includegraphics[width=0.95\linewidth]{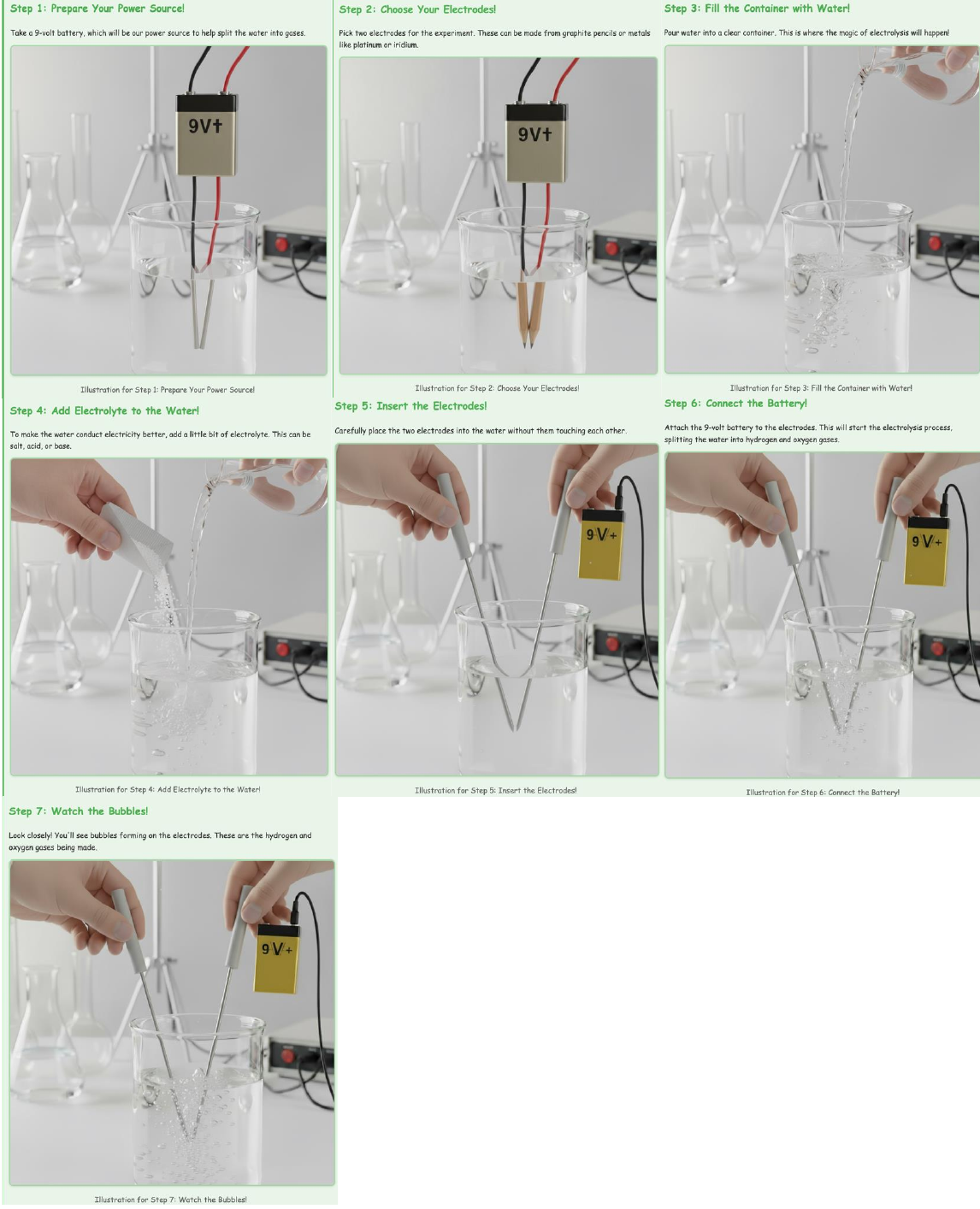}
    \caption[]{Our generated e-booklet (Continued). \\ \textbf{(c) Experiments steps}}
    \label{fig:booklet-c}
\end{figure*}

\clearpage
\begin{figure*}[htbp]
    \ContinuedFloat
    \centering
    \includegraphics[width=0.95\linewidth]{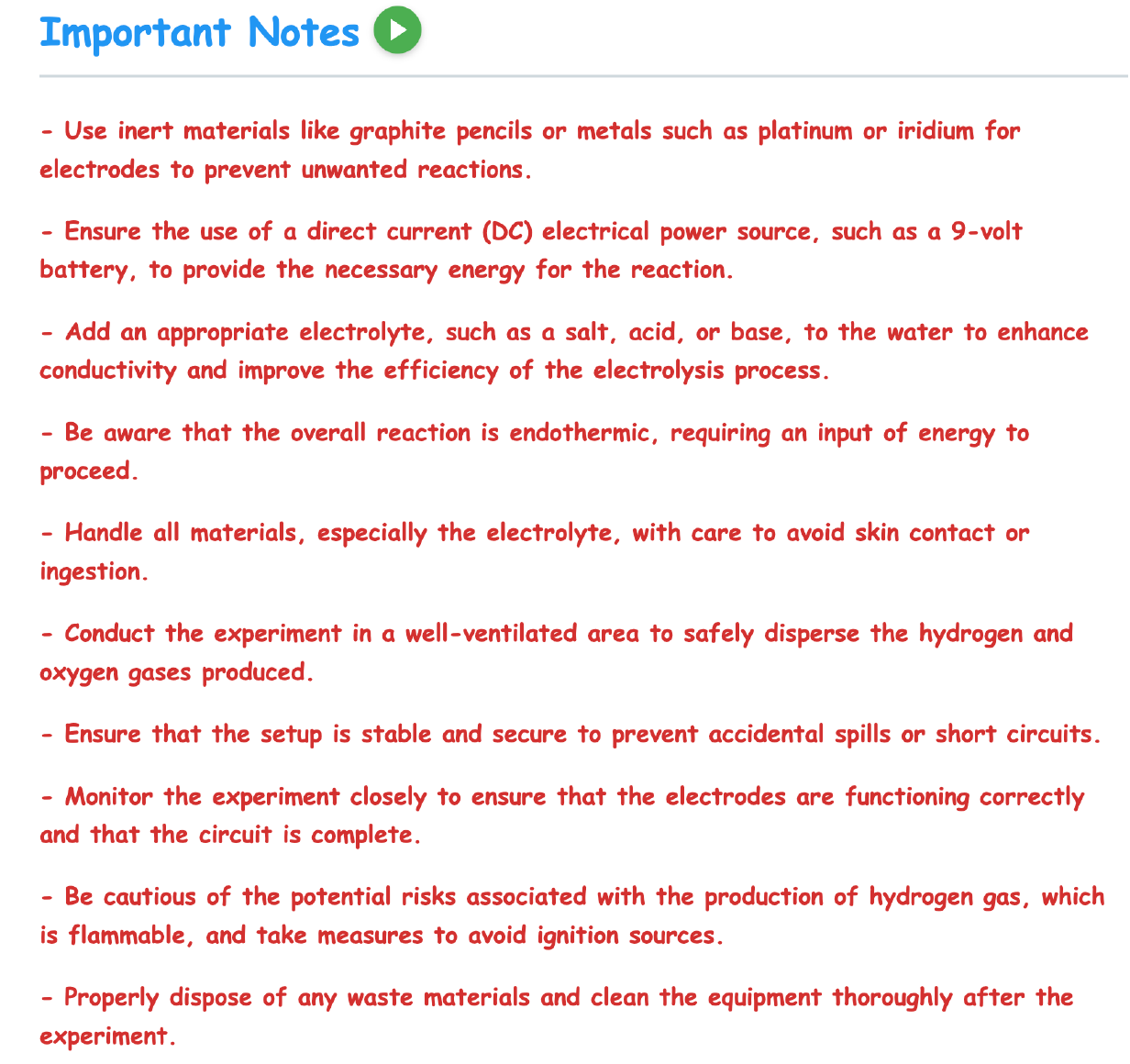}
    \caption[]{Our generated e-booklet (Continued). \\ \textbf{(d) Important Notes}}
    \label{fig:booklet-d}
\end{figure*}

\begin{figure*}[htbp]
    \ContinuedFloat
    \centering
    \includegraphics[width=0.95\linewidth]{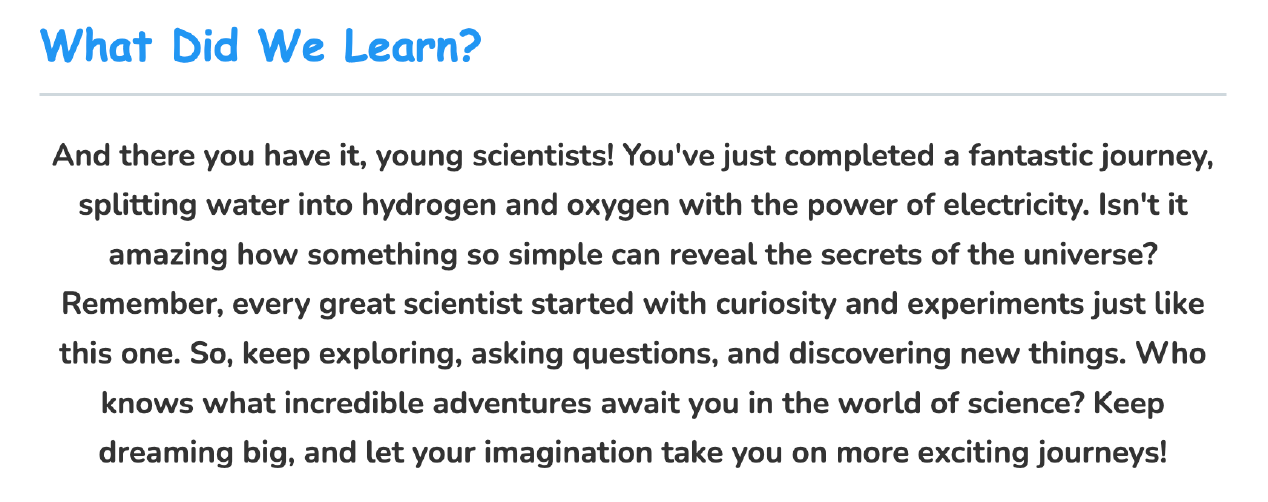}
    \caption[]{Our generated e-booklet (Continued). \\ \textbf{(e) Summary}}
    \label{fig:booklet-e}
\end{figure*}
\clearpage % 结束图片展示，后续正文继续

\section{Extra Information}
\subsection{Knowledge Base}
The knowledge base contains fundamental scientific concepts and detailed explanations in physics and chemistry. For instance, in physics, it covers topics such as Newton's second law, electromagnetic induction, and thermal expansion and contraction. In chemistry, it includes the combustion of metals and the properties of gases in the air. It also incorporates essential physics formulas and chemical equations. The scope of knowledge spans basic science from middle school to high school levels. Beyond this, it strictly excludes any advanced knowledge or uncommon scientific phenomena. The knowledge base is structured into 84 chapters, each with a specific theme. Its primary purpose is to serve as a foundational reference document corpus for IDF (Inverse Document Frequency) value calculation. Additionally, it aims to mitigate hallucinations produced by large language models during the plan stage by providing them with a fundamental knowledge context. We utilize the text-embedding-3-large model to compute and store embeddings for each chapter of the knowledge base. This allows for rapid vector retrieval in subsequent system runs, eliminating the need to recompute the knowledge base document embeddings each time. The concept behind this knowledge base can be applied to other fields as well, and is not limited to the domain of scientific video understanding.

\subsection{Average Cost Per Question}

We measure the average time consumption and token (monetary) cost per question for SciEducator when the maximum number of PDSA Cycle iterations during video understanding is set to 1, 3, and 5. The results are shown below. Note that in addition to the time and tokens consumed by the PDSA Cycle itself, a fixed call to the Captioner Agent is required beforehand to obtain an initial video description.
\begin{itemize}
    \item Maximum PDSA rounds = 1: Average time consumption per question is about 105s, with a money cost of \$0.0542.
    \item Maximum PDSA rounds = 3: Average time consumption per question is about 158s, with a money cost of \$0.0783.
    \item Maximum PDSA rounds = 5: Average time consumption per question is about 206s, with a money cost of \$0.1051.
\end{itemize}

All money costs are automatically calculated by the platform of the APIs we call.

\subsection{SciVBench Dataset}
\begin{figure}[t]
    
    \centering
    \includegraphics[width=1.0\linewidth,height=5cm]{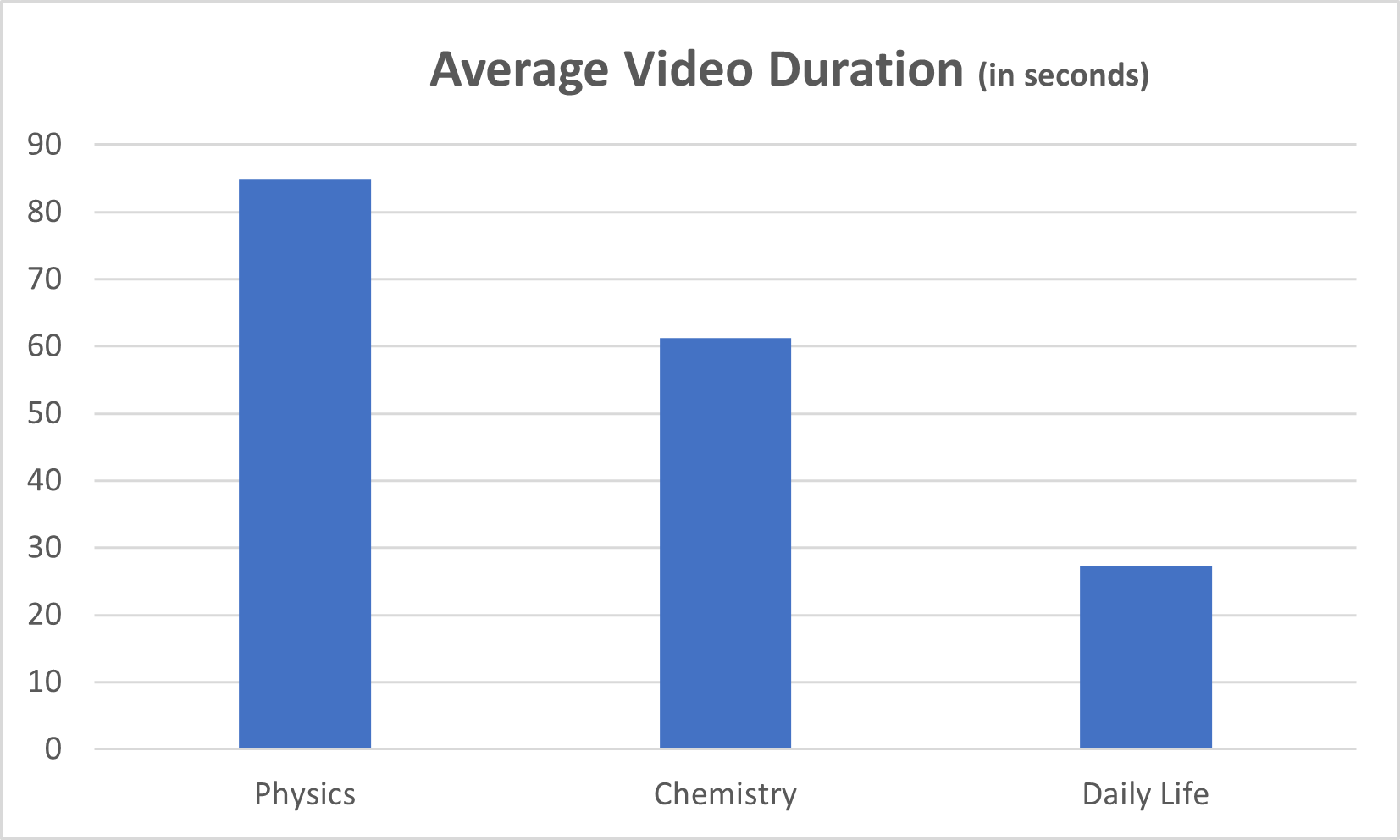}
     \caption[]{Statistics of average duration for three video categories in SciVBench.}
     \label{figure 9}
\end{figure}

\begin{figure}[t]
    
    \centering
    \includegraphics[width=1.0\linewidth,height=5cm]{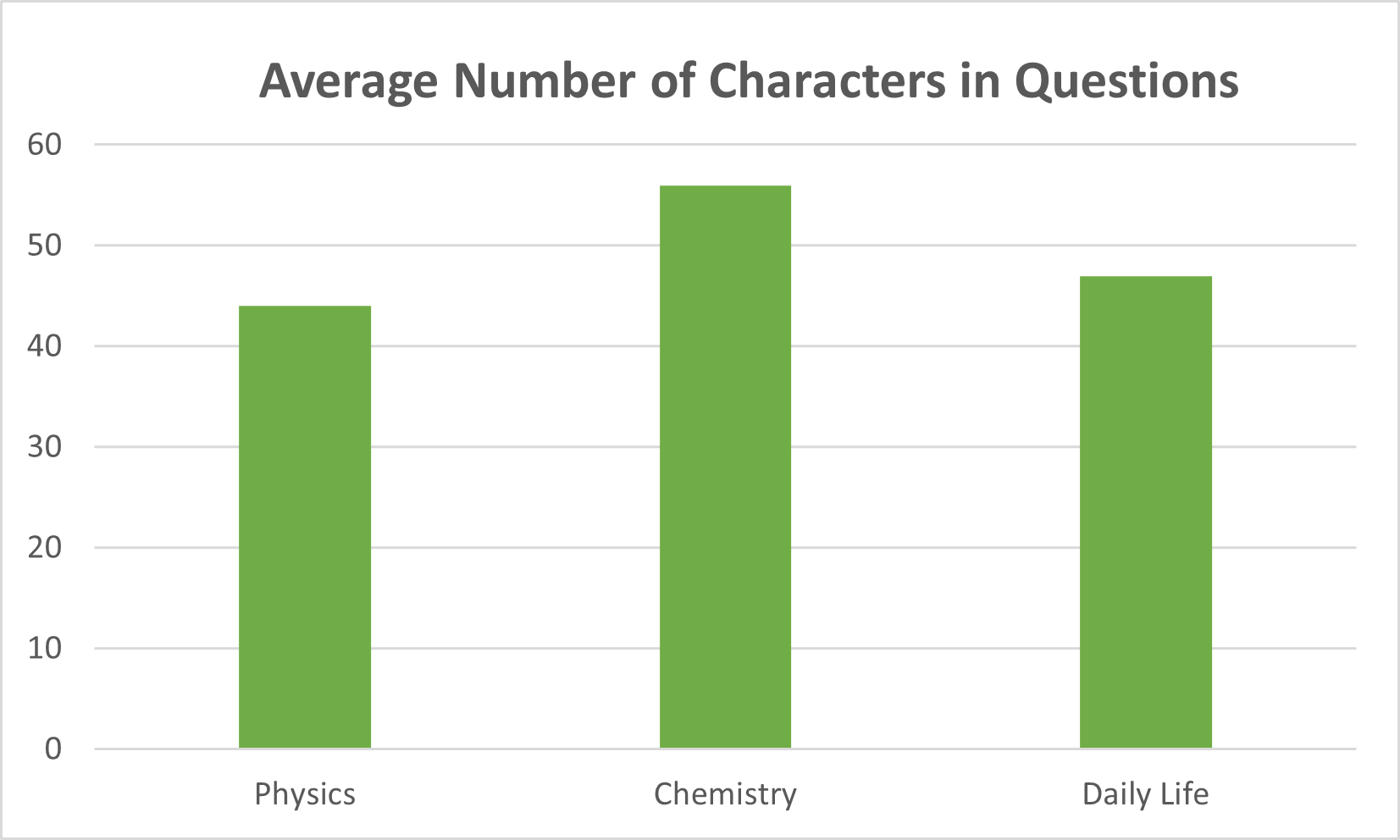}
   \caption[]{Statistics of average question character length for three video QA categories in SciVBench.}
   \label{figure 10}
\end{figure}

\begin{figure}[t]
    
    \centering
    \includegraphics[width=1.0\linewidth,height=5cm]{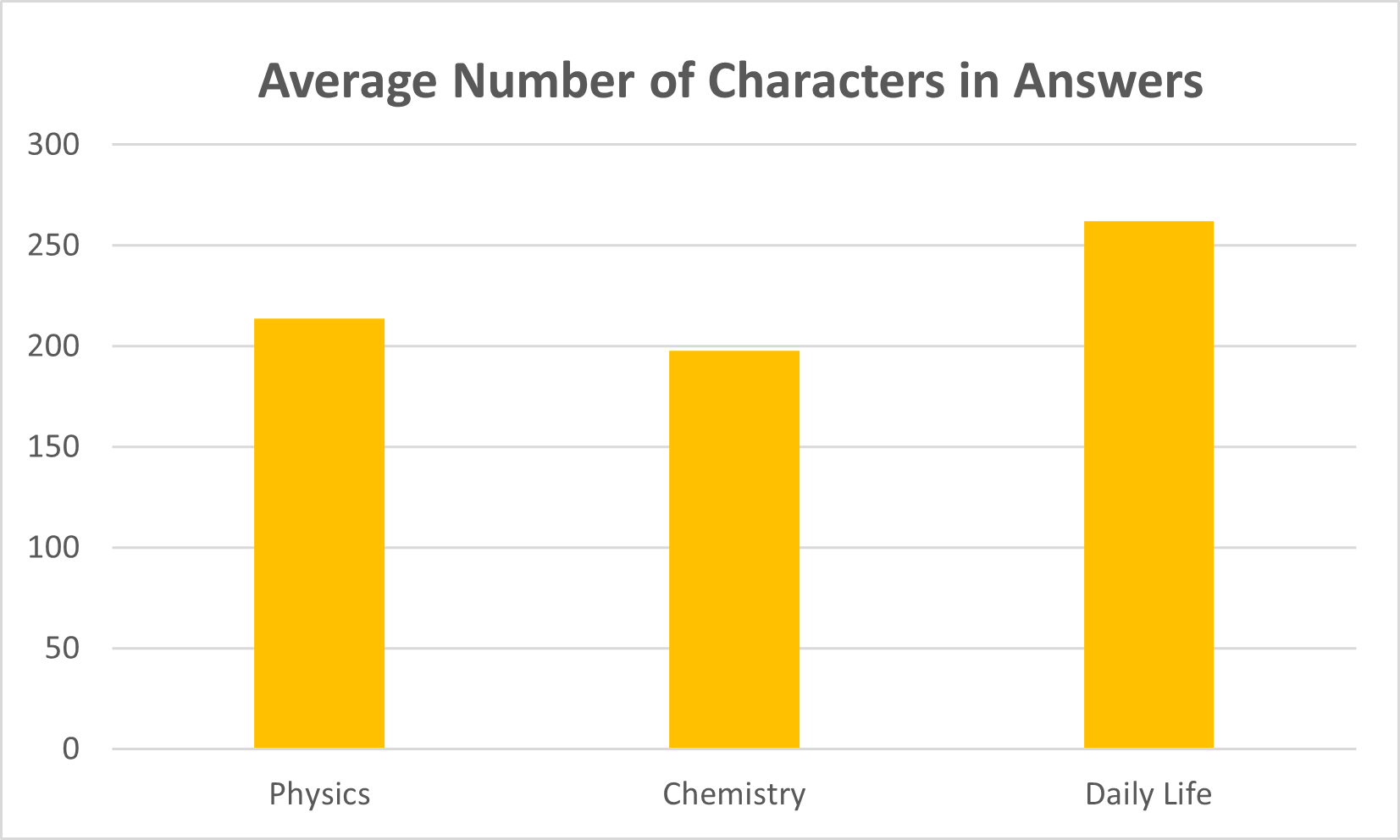}
     \caption[]{Statistics of average answer character length for three video QA categories in SciVBench.}
     \label{figure 11}
\end{figure}

We provide some statistical analyses of videos and QA pairs in SciVBench. ~\cref{figure 9} presents statistics on the average video duration. ~\cref{figure 10} and ~\cref{figure 11} show statistical analyses of question and answer lengths. All three figures are presented separately for physics, chemistry, and daily life.

